\documentclass[sigconf, screen, nonacm]{acmart}

\AtBeginDocument{%
  \providecommand\BibTeX{{%
    \normalfont B\kern-0.5em{\scshape i\kern-0.25em b}\kern-0.8em\TeX}}}

\usepackage{subfigure}

\begin{document}

\title[AutoDRIVE Simulator]{AutoDRIVE Simulator: A Simulator for Scaled Autonomous Vehicle Research and Education}


\author{Tanmay Samak and Chinmay Samak}
\authornote{Both authors contributed equally to this research. This work is an outcome of India Connect@NTU Research Internship Programme, 2020.}
\email{{tv4813, cv4703}@srmist.edu.in}
\affiliation{%
  \institution{SRM Institute of Science and Technology}
  \city{Kattankulathur}
  \state{Tamil Nadu}
  \country{India}
  \postcode{603203}
}

\author{Ming Xie}
\email{mmxie@ntu.edu.sg}
\affiliation{%
  \institution{Nanyang Technological University}
  \city{50 Nanyang Avenue}
  \country{Singapore}
  \postcode{639798}
}


\begin{abstract}
	AutoDRIVE is envisioned to be an integrated research and education platform for scaled autonomous vehicles and related applications. This work is a stepping-stone towards achieving the greater goal of realizing such a platform. Particularly, this work introduces the AutoDRIVE Simulator, a high-fidelity simulator for scaled autonomous vehicles. The proposed simulation ecosystem is developed atop the Unity game engine, and exploits its features in order to simulate realistic system dynamics and render photorealistic graphics. It comprises of a scaled vehicle model equipped with a comprehensive sensor suite for redundant perception, a set of actuators for constrained motion control and a fully functional lighting system for illumination and signaling. It also provides a modular environment development kit, which comprises of various environment modules that aid in reconfigurable construction of the scene. Additionally, the simulator features a communication bridge in order to extend an interface to the autonomous driving software stack developed independently by the users. This work describes some of the prominent components of this simulation system along with some key features that it has to offer in order to accelerate education and research aimed at autonomous driving.
\end{abstract}



\begin{teaserfigure}
  \includegraphics[width=\textwidth]{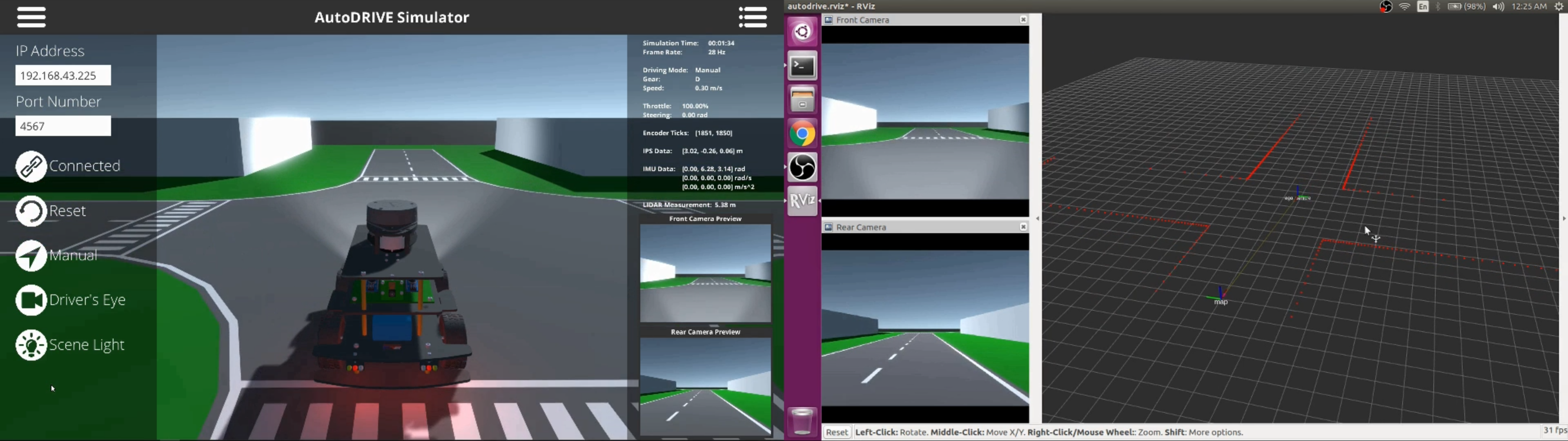}
  \caption{AutoDRIVE Simulator interfaced with Robot Operating System in a distributed computing configuration.}
  \Description{AutoDRIVE Simulator running on a Windows machine, depicted in the left-half, is interfaced with the Robot Operating System (ROS) running on an Ubuntu machine through a WebSocket interface, and the real-time sensor data is being visualized in ROS Visualization (RViz) tool; depicted in the right-half.}
  \label{fig:Teaser}
\end{teaserfigure}

\maketitle

\section{Introduction}

The field of autonomous driving has witnessed an accelerated advancement over the past few decades owing to the continuous and focused research. Furthering this technology demands rapid development and testing of various aspects of vehicle autonomy through novel algorithms. The budding engineers must also be educated well about this technology in order to prepare them to work alongside the experts in the field. Both of these tasks in turn call for design and development of efficient tools aiding researches and educators throughout the process.

Considering the monetary and spatial constraints for most of the students and self-funded labs, a scaled platform for autonomous vehicle research and education is one of the most viable option. However, the present autonomous driving research community lacks integrated tools for flexible design and development of autonomy algorithms targeted specifically towards scaled autonomous vehicles. Most researchers tend to employ simulators targeted towards full-scale autonomous vehicles, but make use of scaled robotic vehicle(s) for real-world deployment. Now, while it may seem fine to transfer the same autonomy algorithms (sim2real), the drastic variation in perception data and system dynamics poses several challenges, some of which may not even be apparent enough.

This work presents AutoDRIVE Simulator, a cross-platform simulator for scaled autonomous vehicles. The simulator is developed atop the Unity game engine, which enables simulation of realistic physics (using NVIDIA’s PhysX engine) and rendering of photorealistic graphics (using Unity’s Post-Processing Stack). The simulator can be exploited by the users (particularly targeting students and researchers in the field) in order to develop and test their algorithms aimed at autonomous driving, before deploying them on a similar real-world scaled autonomous vehicle.

The scope of the presented simulation system includes design and development of a scaled vehicle (with realistic sensor and actuator models, accurate system dynamics and fully functional lighting system), several environment modules, a communication bridge and a convenient user interface. It is to be noted that development of the autonomy algorithms is left to the users, and that sufficient resources are provided for this purpose.

\section{Related Work}

Simulators are a key value addition to the research phase as they allow virtual prototyping of the system and surrounding under study. They ensure physical safety and enable rapid prototyping at minimal costs. They also allow simulation of critical situations and corner-cases which are rarely encountered in real-world along with permutations and combinations of various sub-scenarios. Furthermore, reproduction of scenario-specific implementations is also possible in a simulated environment.

The automotive industry has long practiced the use of simulators to simulate near-realistic vehicle dynamics, specifically for rapid design and development of automotive sub-systems. Ansys Automotive \cite{AnsysAutomotive} and Adams Car \cite{AdamsCar} are prime examples of commercial simulators adopted by industry and academia for simulating realistic dynamics for vehicle design and testing. However, with the advent of autonomous vehicles, advanced driver assistant systems (ADAS) in particular, most of these simulators have started releasing updates that include autonomous driving features of some form or the other. Simulators such as Ansys Autonomy \cite{AnsysAutonomy}, CarMaker \cite{CarMaker} and CarSim \cite{CarSim} are few amongst a wide range of such automotive simulators that have released autonomous driving updates in the recent past.

TORCS \cite{TORCS} is one of the earliest open-source simulators, which focuses predominantly on the task of manual as well as autonomous racing. Gazebo \cite{Gazebo} is one of the most popular open source simulators for robotics research and development, and has been natively adopted by the Robot Operating System (ROS) \cite{ROS}. AirSim \cite{AirSim}, CARLA \cite{CARLA}, LGSVL Simulator \cite{LGSVLSimulator} and Deepdrive \cite{Deepdrive} are some of the other open source simulators that particularly focus on the problem of autonomous driving.

Commercial automotive simulators focusing on autonomous driving and driver assistance systems include NVIDIA Drive Constellation \cite{DRIVEConstellation}, rFpro \cite{rFpro}, dSPACE \cite{dSPACE}, PreScan \cite{PreScan}, Cognata \cite{Cognata} and Metamoto \cite{Metamoto}. Researchers have also adopted commercial video games such as Grand Theft Auto V \cite{Richter2016, Richter2017, Johnson-Roberson2017} to simulate autonomous vehicles.

However, none of these simulators support scaled autonomous vehicles and environments at full capacity, which was the key problem statement for this research project.

\section{Overview}

\begin{figure}[h]
	\centering
	\includegraphics[width=\linewidth]{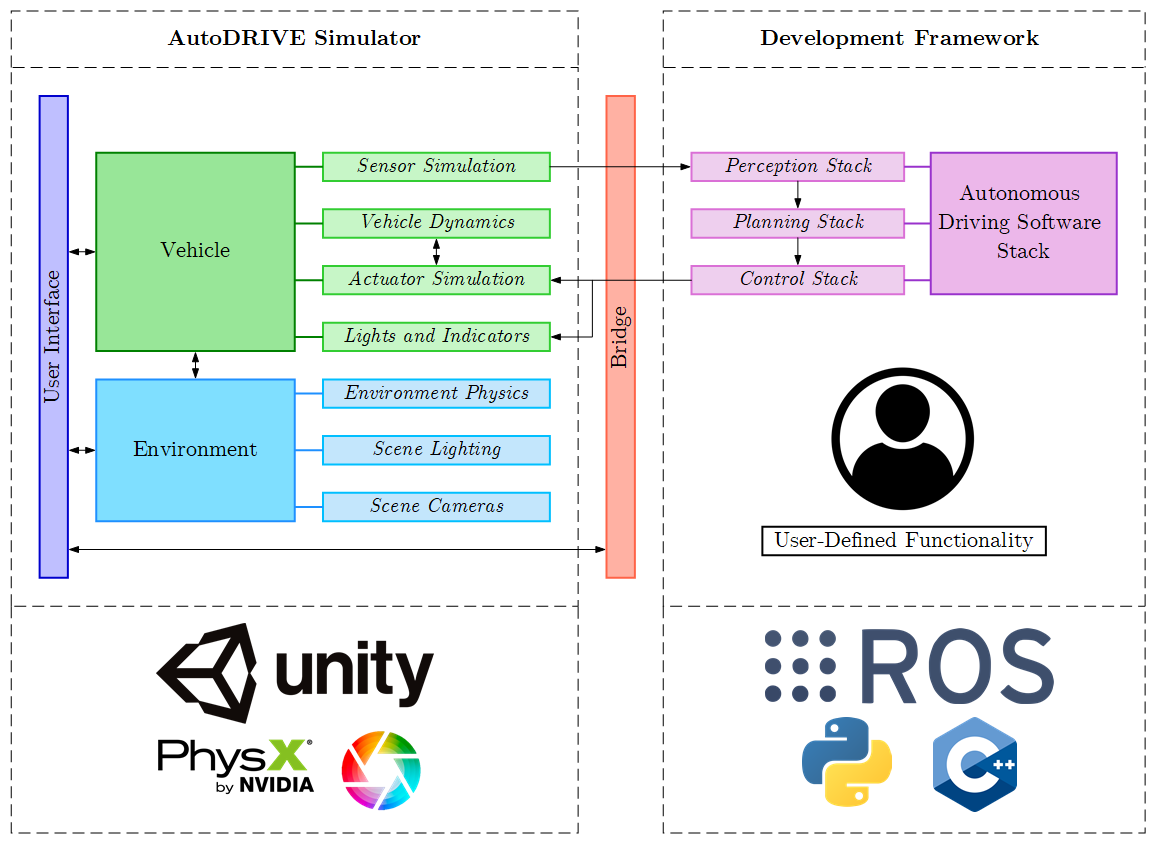}
	\caption{Overview of the proposed system.}
	\Description{The AutoDRIVE Simulator comprises of a scaled vehicle and a modular environment. It also features a user interface and a communication bridge. The development framework comprises of necessary tools for developing the autonomous driving software stack.}
	\label{fig:Overview}
\end{figure}

Figure \ref{fig:Overview} illustrates overview of the proposed system, and distinctly depicts components of the simulator and development framework. While the vehicle, environment and user interface strictly fall under the domain of simulator development, the communication bridge is incorporated within both, the simulator as well as the development framework, in order to interface the two with each other.

The task of simulator development was split into four distinct stages, viz. design and development of the vehicle, the environment, the communication bridge and the user interface. Specifically, vehicle development comprised of designing the vehicle mesh model, modelling realistic sensing modalities and actuators, definition of accurate system dynamics, and provision of a fully functional lighting system. On the other hand, the environment development comprised of designing various environment modules and a preconfigured map, definition of realistic environment physics and scene lighting properties, and provision of multiple scene cameras.

Additionally, a development framework was to be designed in order to provide the users with the necessary tools for developing and interfacing autonomy algorithms with the simulator. It comprises of a ROS package along with Application Programming Interfaces (APIs) in order to provide direct scripting support for Python and C++. As discussed earlier, the task of implementing autonomous driving software stack is outside the scope of this research project and is left to the users.

\section{AutoDRIVE Simulator}

The AutoDRIVE Simulator is developed atop the Unity game engine and exploits its built-in multi-threaded physics simulation engine called PhysX for simulating realistic system dynamics. It also utilizes the Post-Processing Stack for rendering photorealistic graphics. The following sections brief about some of its prominent components.

\subsection{Vehicle}

The vehicle was designed with an aim of enabling hardware development in the future, thereby achieving the second milestone in the development of AutoDRIVE Platform.

\begin{figure}[h]
	\centering
	\includegraphics[width=\linewidth]{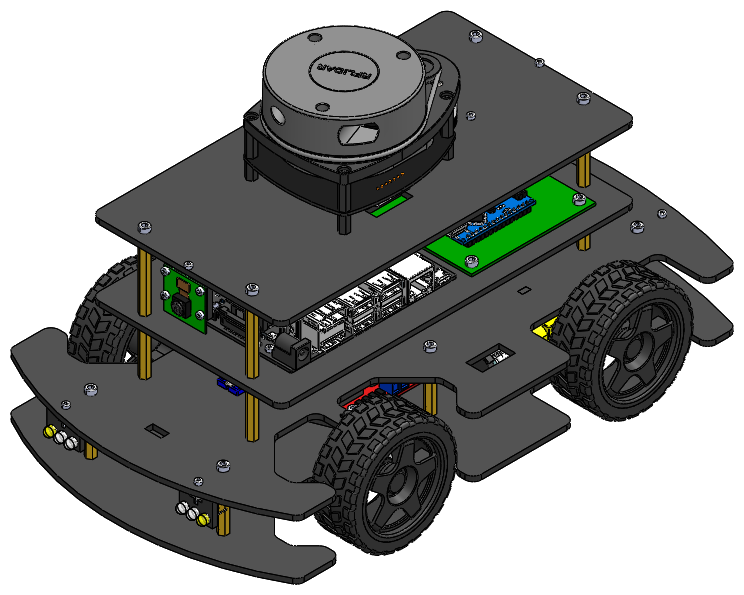}
	\caption{3D model of the scaled vehicle.}
	\Description{3D Computer-Aided Design (CAD) of the scaled vehicle developed using SolidWorks.}
	\label{fig:VehicleModel}
\end{figure}

The designed vehicle (Figure \ref{fig:VehicleModel}) is a 1:14 scale front-wheel steered, rear-wheel drive model. It primarily comprises of four different platforms fastened together at appropriate offsets using standoffs. The first platform mounts power electronics components, drive actuators and steering mechanism. The second platform holds steering actuator coupled to the steering link and also serves as a place to strap on a LiPo battery pack. It also has provision for mounting front and rear light modules. The third platform accommodates an on-board computer and a custom circuit board with power distribution circuitry, low-level microcontrollers, sensors and other interfacing components. Finally, the fourth platform houses a Light Detection and Ranging (LIDAR) unit and also mounts the front and rear-view cameras.

\noindent{\textbf{\\Sensor Simulation:\\}}

The simulator currently supports seven different sensors, viz. throttle sensor, steering sensor, incremental encoders, Indoor Positioning System (IPS), Inertial Measurement Unit (IMU), LIDAR and cameras.

\begin{itemize}
	
	\item \textbf{Throttle sensor:} The throttle sensor is a virtual sensor, which measures the instantaneous throttle value of the vehicle in the range of [-1, 1]. While positive readings of the throttle sensor indicate forward driving, negative readings indicate reverse driving. A zero-throttle reading implies that the vehicle is in a stand-still condition as the brakes are automatically applied.
	
	\item \textbf{Steering sensor:} The steering sensor is a virtual sensor, which measures the instantaneous steering angle of the vehicle in the range of [-1, 1]. While positive readings of the steering sensor indicate right turns, negative readings indicate left turns. A zero-steering reading implies that the vehicle is driving straight.
	
	\item \textbf{Incremental encoders:} The drive actuators are each installed with a simulated incremental encoder. These encoders have a resolution of 16 Pulses Per Revolution (PPR) and measure the angular displacement of the motor shafts (and hence the wheels, after appropriate gearbox magnification) and update the count once the wheels have turned by a predefined amount. They output the pulse count (ticks) and angle turned by the wheels.
	
	\item \textbf{IPS:} The simulated IPS is a scaled form of Global Navigation Satellite System (GNSS). It localizes vehicle within the map and provides the position vector as output.
	
	\item \textbf{IMU:} The simulated 9 Degrees of Freedom (DoF) IMU measures inertial data of the vehicle. This includes the vehicle’s orientation (both quaternion and Euler angles are supported), angular velocity and linear acceleration. The inertial readings are converted from Unity’s native left-handed co-ordinate system into a right-handed co-ordinate system with X-axis pointing to the front of the vehicle, Y-axis pointing to the left of the vehicle and Z-axis pointing to the top of the vehicle.
	
	\item \textbf{LIDAR:} The simulated LIDAR sensor measures the relative distance of objects in the scene using ray-casting. The simulator implements a 2D LIDAR, which performs a 360° scan around the vehicle at 7 Hz update rate, with an angular resolution of 1°, a minimum range of 0.15 m and maximum range of 12 m. The range array is provided as output along with a simulated intensity array.
	
	\item \textbf{Cameras:} The simulated vehicle has front and rear-view cameras, which provide visual sensory data in the form of RGB image frames. The cameras are simulated with a 41.669° Field of View (FoV), a sensor size of 3.76$\times$2.74 mm, a focal length of 3.6 mm and a target resolution of 640$\times$480p.
	
\end{itemize}

\noindent{\textbf{\\Vehicle Dynamics:\\}}

The notion of physical modelling was adopted in order to develop the dynamic model of the vehicle, and Unity's inbuilt PhysX engine was used for real time computation and updation of the vehicle motion based on the control inputs and its resulting interaction with the environment.

\begin{figure}[h]
	\centering
	\includegraphics[width=\linewidth]{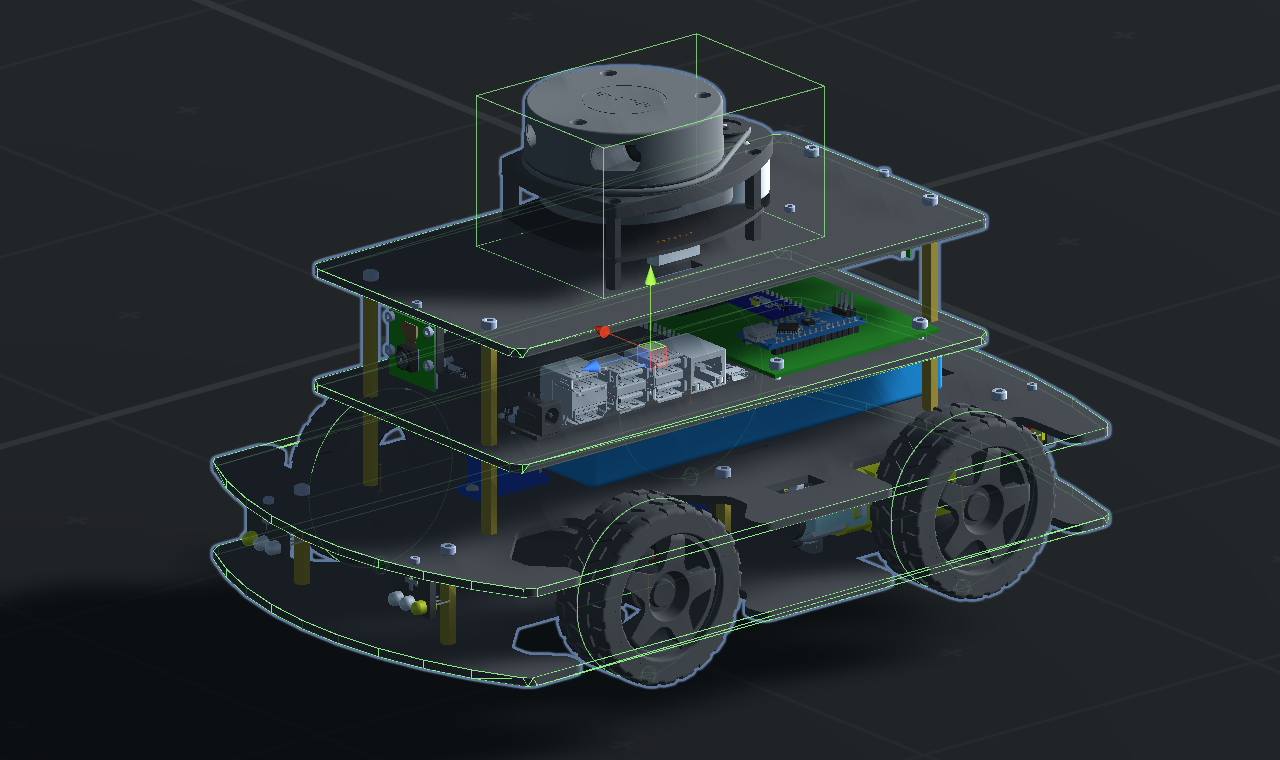}
	\caption{Physical model of the scaled vehicle.}
	\Description{Physical model of the scaled vehicle in Unity Scene comprising of rigid body, mesh colliders (for the four platforms), box collider (for the LIDAR unit) and wheel colliders.}
	\label{fig:VehicleDynamics}
\end{figure}

The developed vehicle model has various physical components associated with it (Figure \ref{fig:VehicleDynamics}), such as a rigid body to simulate gravity and other forces acting on the vehicle, body colliders to simulate realistic collisions and, most importantly, wheel colliders to simulate realistic driving behavior. In addition to this, all the system parameters such as mass, air drag, longitudinal and lateral friction, etc. are configured to simulate realistic vehicular dynamics.

\begin{figure*}[h]
	\centering
	\subfigure[]{\includegraphics[width=0.33\textwidth]{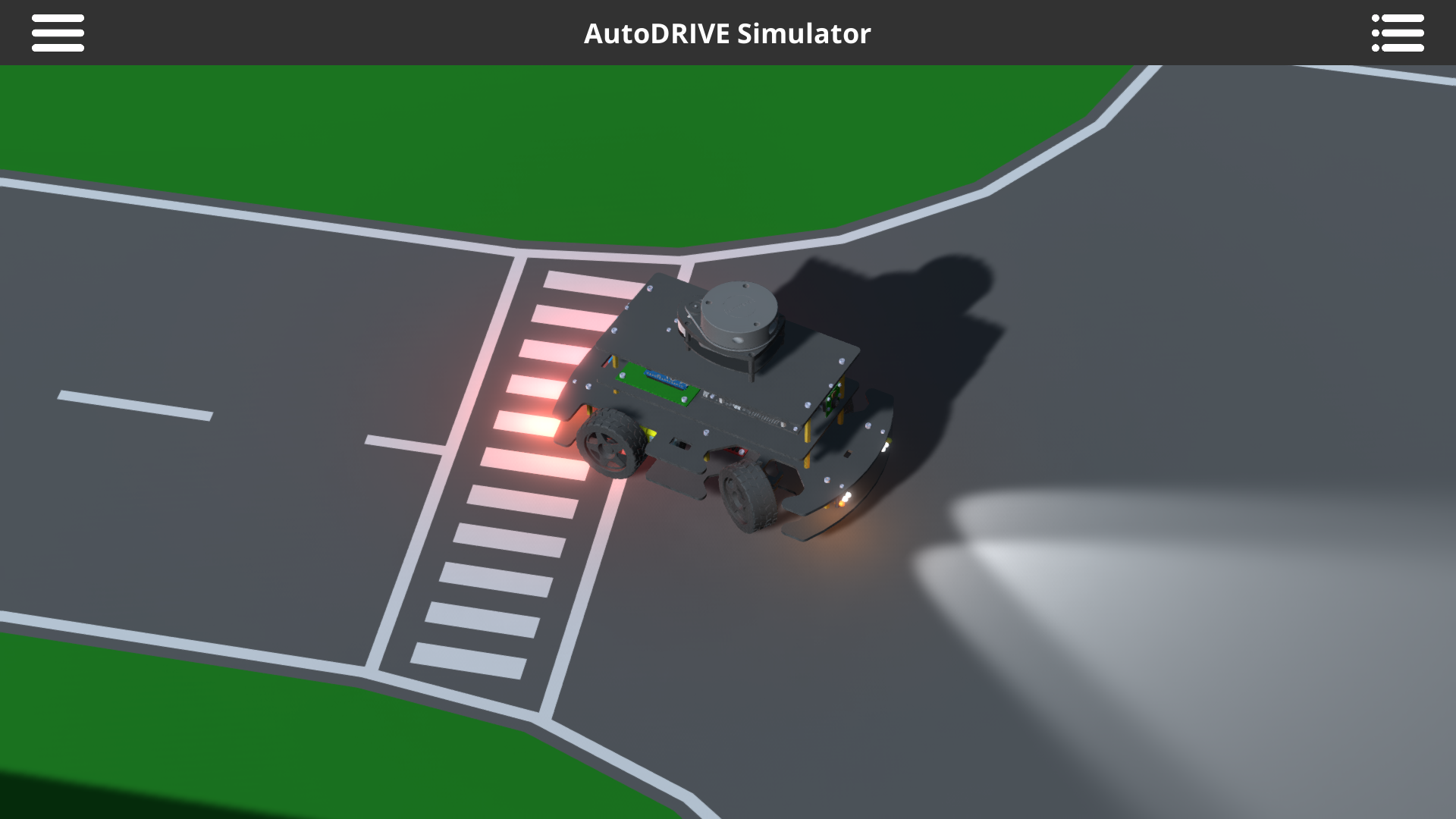}}
	\subfigure[]{\includegraphics[width=0.33\textwidth]{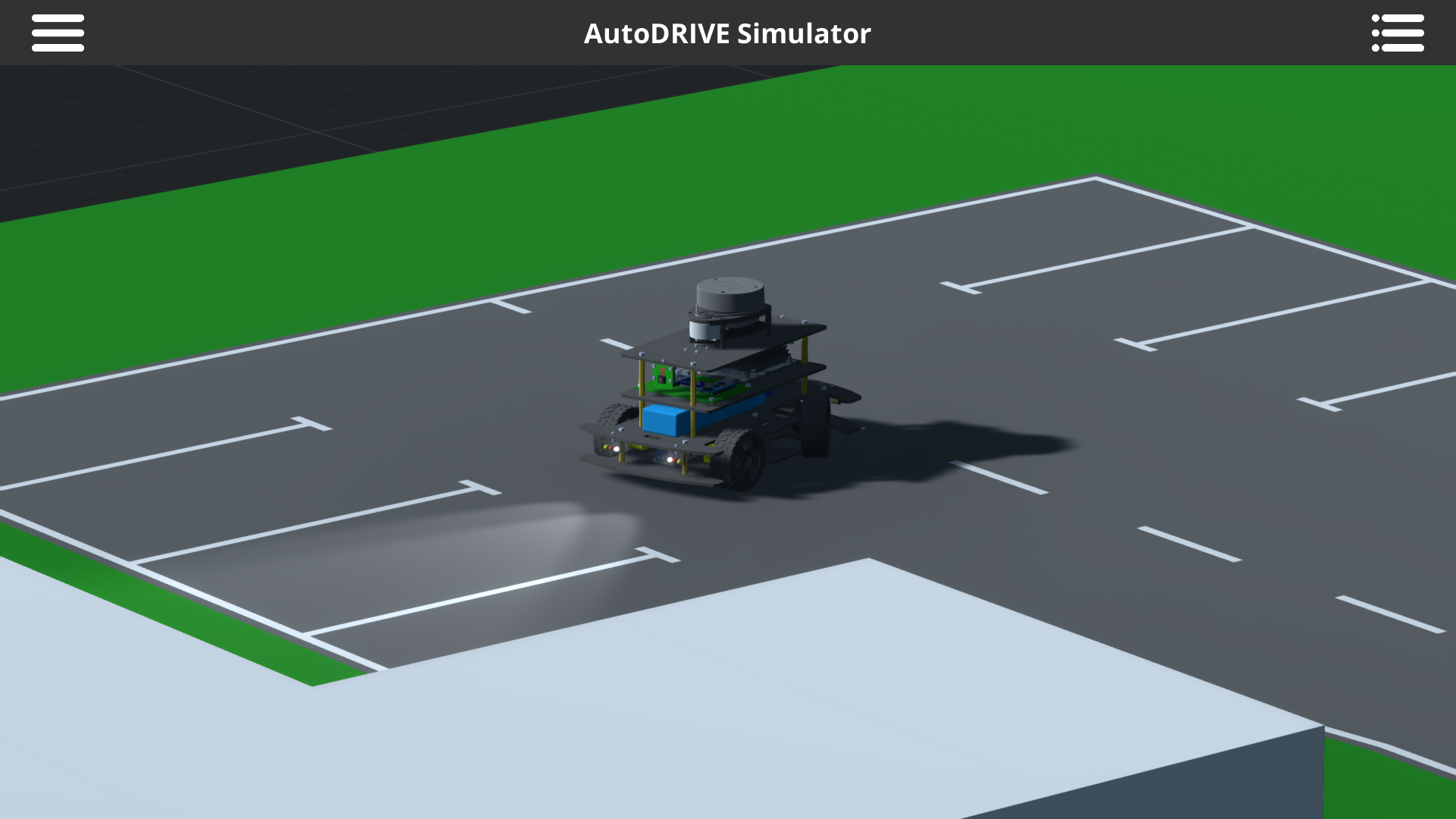}}
	\subfigure[]{\includegraphics[width=0.33\textwidth]{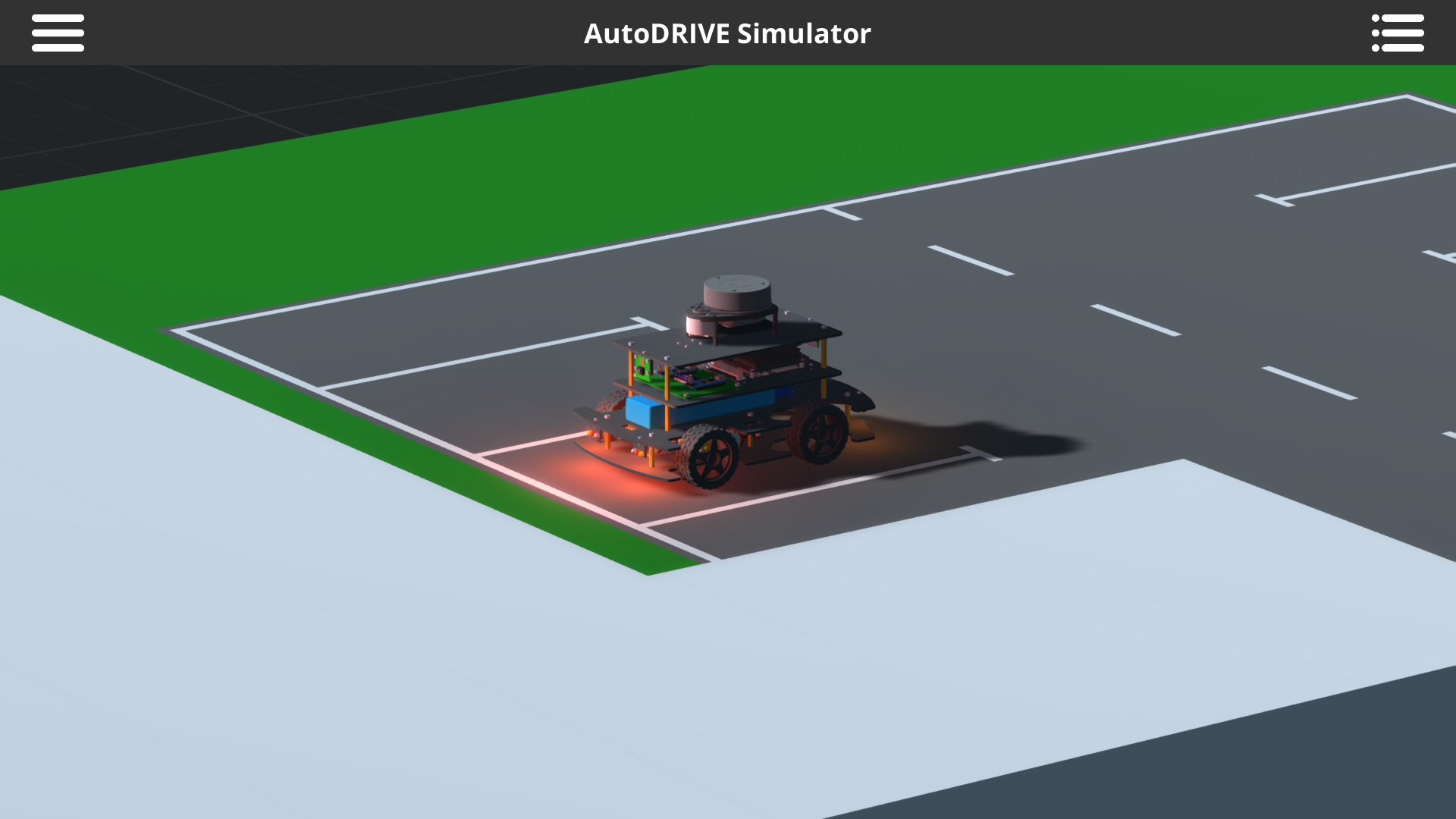}}
	\caption{Vehicle lighting demonstrations: (a) headlights, taillights and right indicators, (b) reverse indicators, and (c) brake and hazard indicators.}
	\Description{Vehicle lighting demonstrations: (a) Vehicle taking right turn at a 4-way intersection has its headlights (low-beam), taillights (partially luminous) and right indicators enabled; (b) Vehicle executing a parking maneuver has its reverse indicators enabled; (c) Parked vehicle has its brake and hazard indicators enabled.}
	\label{fig:VehicleLighting}
\end{figure*}

\noindent{\textbf{Actuator Simulation:\\}}

The vehicle actuation system comprises of two drive actuators (for longitudinal motion control) and a steering actuator (for lateral motion control). While the drive actuators control the torque being applied to the rear wheels based on the specified throttle command, the steering actuator steers the front wheels through a specified steering angle.

In order to simulate the high holding torque of the drive actuators, brakes are automatically applied unless the throttle input is non-zero. Additionally, the steering actuator is retracted to 0° unless any valid steering control input is given to the vehicle.

The actuators accept normalized control inputs so as to remove their dependency over the vehicle geometry and dynamics. All the actuators have been defined with realistic saturation limits (steering actuation limit of ±30° and drive actuation limit of 130 RPM). Furthermore, actuator response delays have also been modelled in order to match realistic actuator dynamics.

\noindent{\textbf{\\Lights and Indicators:\\}}

As depicted in Figure \ref{fig:VehicleLighting}, the vehicle has fully functional headlights (disabled, low-beam, high-beam), taillights/brake indicators (disabled, partially luminous when headlights are illuminated, bright when brakes are applied), turning/hazard indicators (disabled, left indicators, right indicators, hazard indicators) and reverse indicators (disabled in drive gear, enabled in reverse gear). While headlights and turning indicators can be controlled in manual as well as autonomous mode, brake and reverse indicators are controlled automatically upon respective event detection.

\subsection{Environment}

AutoDRIVE is specifically targeted towards scaled autonomous vehicles. As a result, the environment is also scaled down appropriately so as to match physical dimensions of the vehicle. The central idea of environment design was to have a finite number of modules, enabling modular and reconfigurable construction of the scene, where each module be classified as drivable or non-drivable segment w.r.t. the vehicle.

Prior to designing the road modules, minimum road curvature had to be defined. This required knowledge of turning radius of the vehicle, which was calculated based on its kinematic constraints, using the following relation:

\begin{displaymath}
R =  \frac{L}{tan(\delta)}
\end{displaymath}
where $R$ is the turning radius, $L$ is the distance between two reference points that need to follow a circular turn, and $\delta$ is the corresponding steering angle. Considering a total chassis length of 300 mm and a steering actuation limit of ±30°, the turning radius of the vehicle chassis comes out to be 600 mm, which was defined as the minimum road curvature. This ensured that the entire vehicle would stay within a particular lane while taking a turn.

The simulator currently supports dual-lane roads with white lane markings (solid at road boundaries and dashed at lane separation), right-angle turns, 3-way and 4-way intersections (with stop-lines and pedestrian crossings marked in white color), green lawn representing non-drivable segments, and a modular construction box to build 3D obstructions.

\begin{figure}[h]
	\centering
	\subfigure[]{\includegraphics[width=0.156\textwidth]{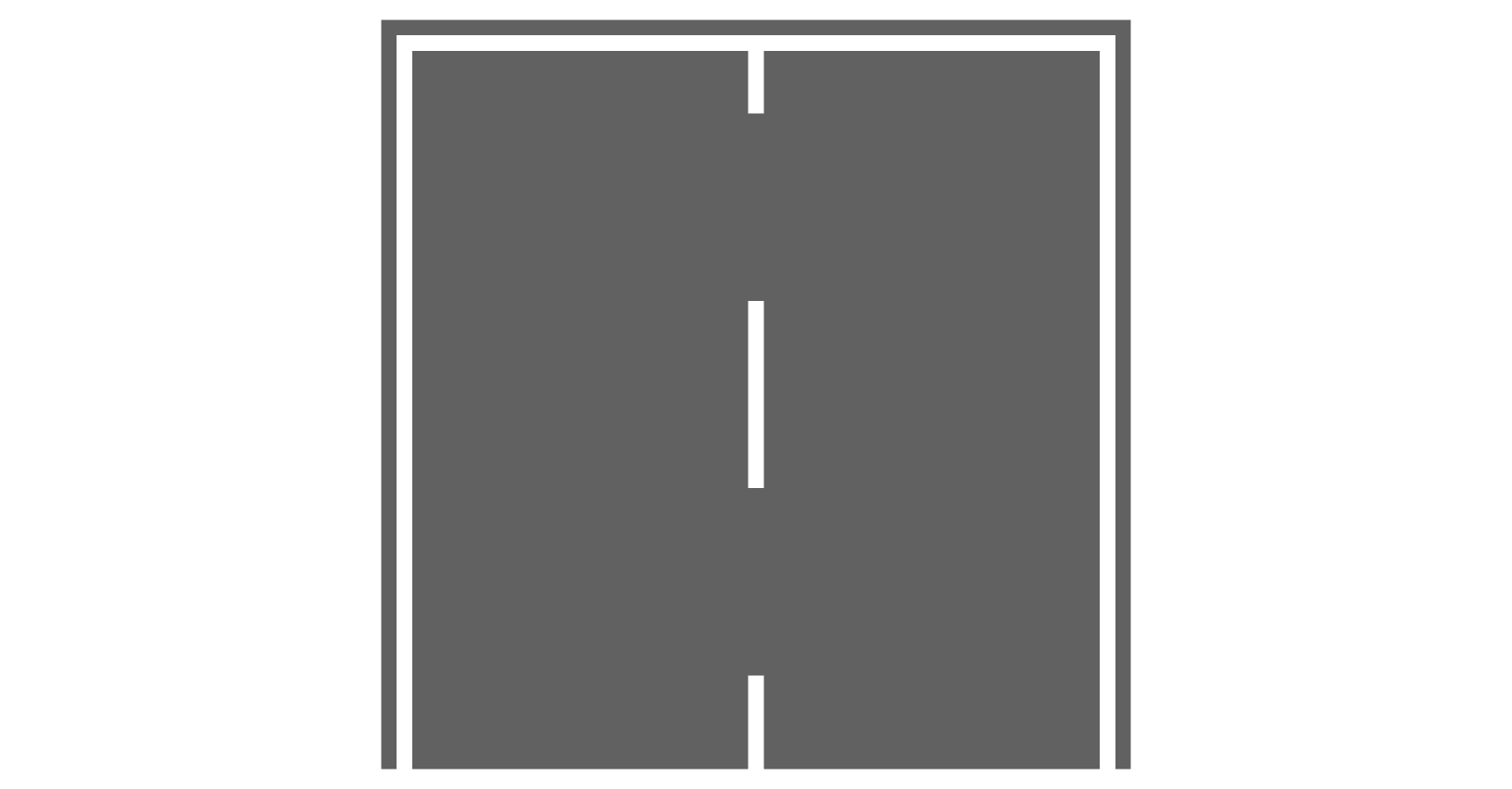}}
	\subfigure[]{\includegraphics[width=0.156\textwidth]{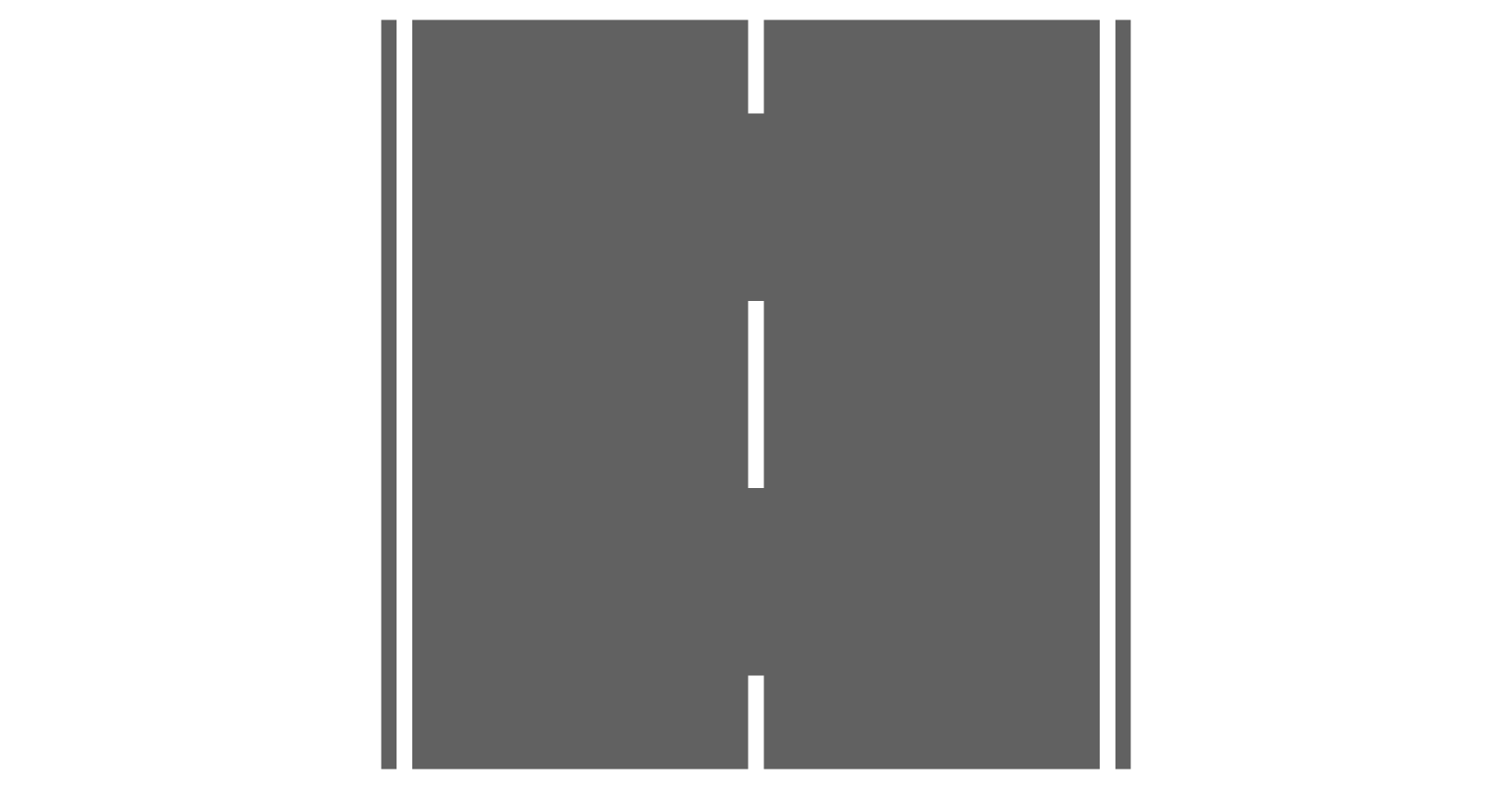}}
	\subfigure[]{\includegraphics[width=0.156\textwidth]{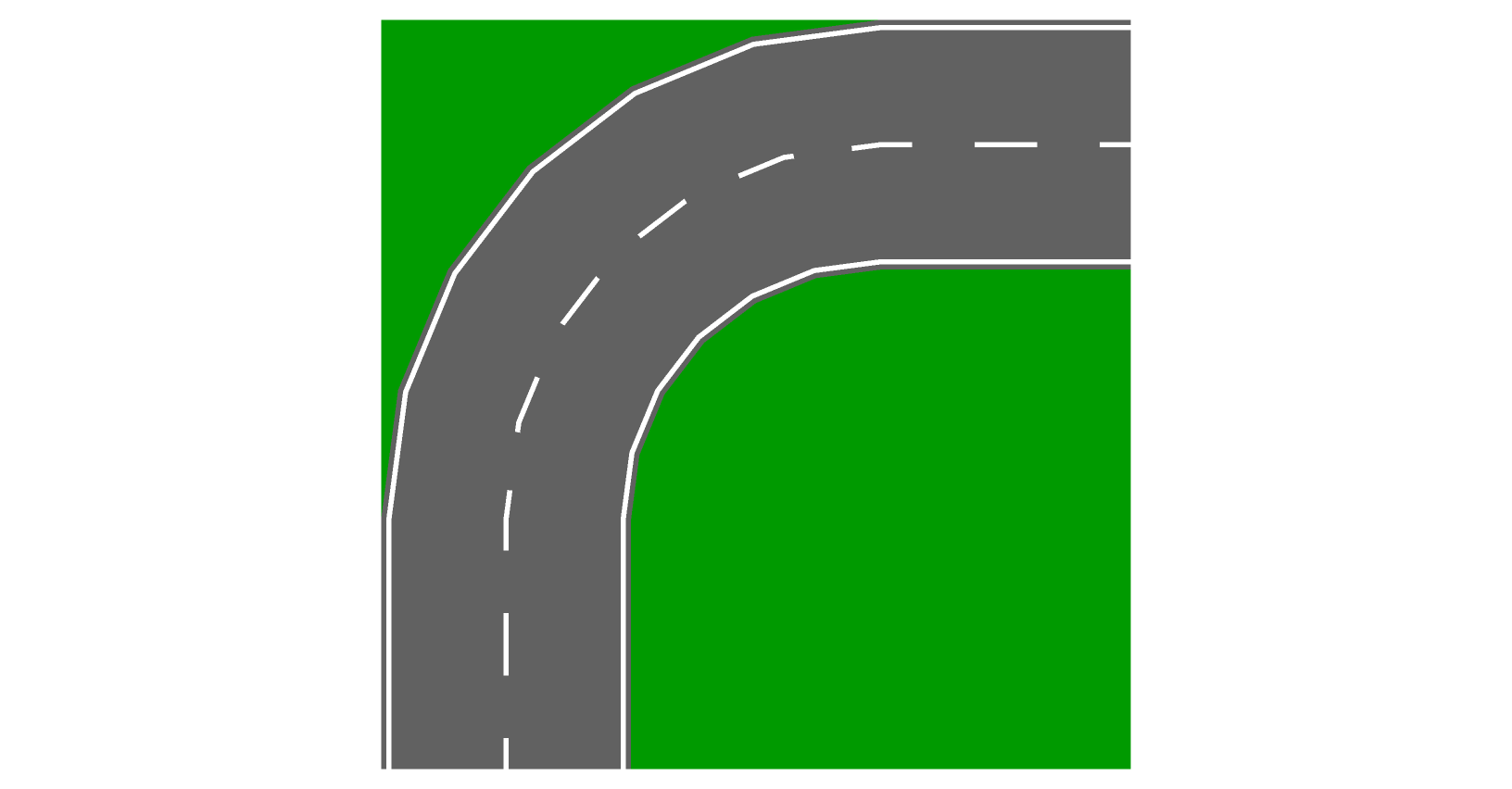}}
	\subfigure[]{\includegraphics[width=0.156\textwidth]{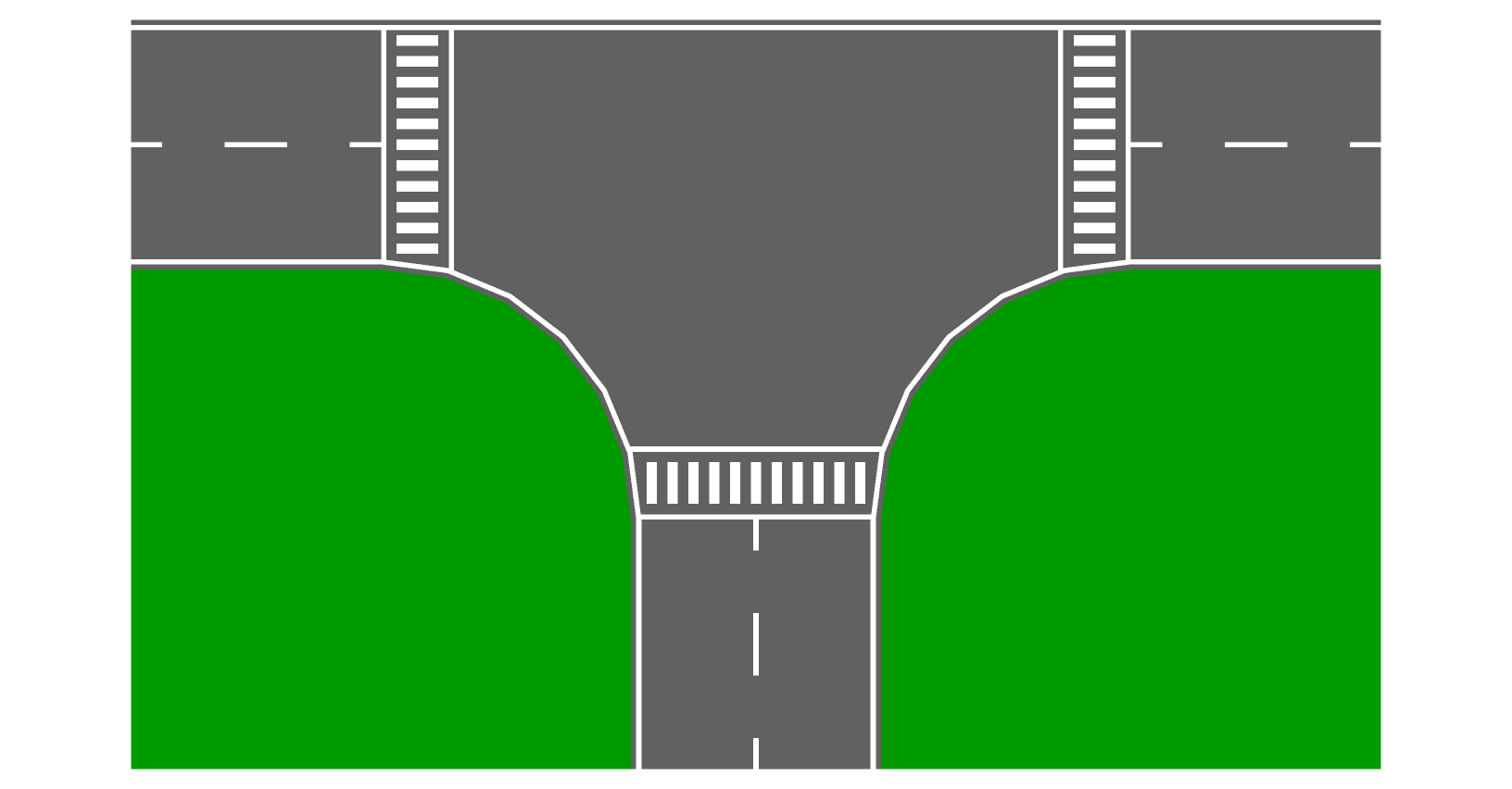}}
	\subfigure[]{\includegraphics[width=0.156\textwidth]{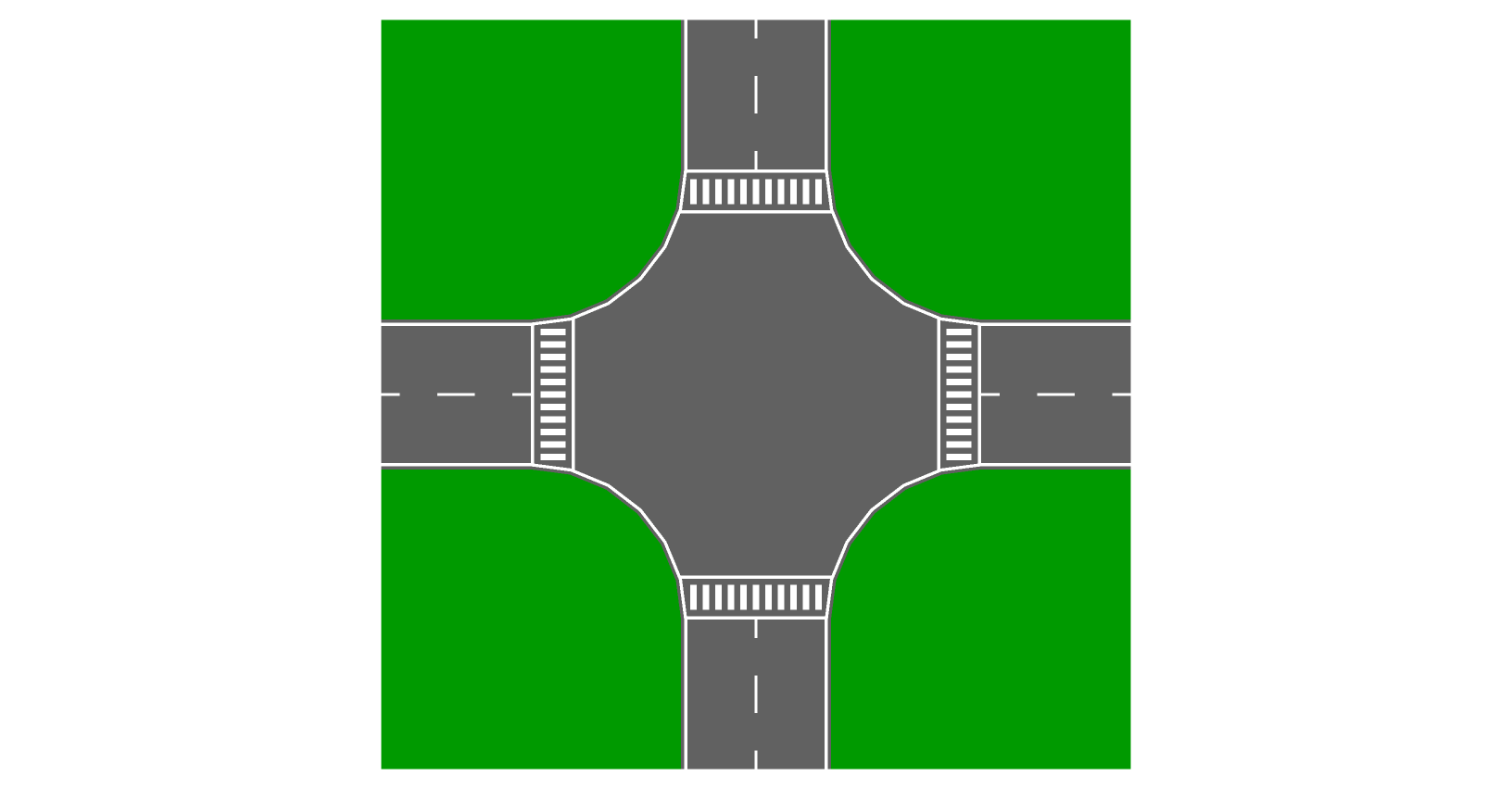}}
	\subfigure[]{\includegraphics[width=0.156\textwidth]{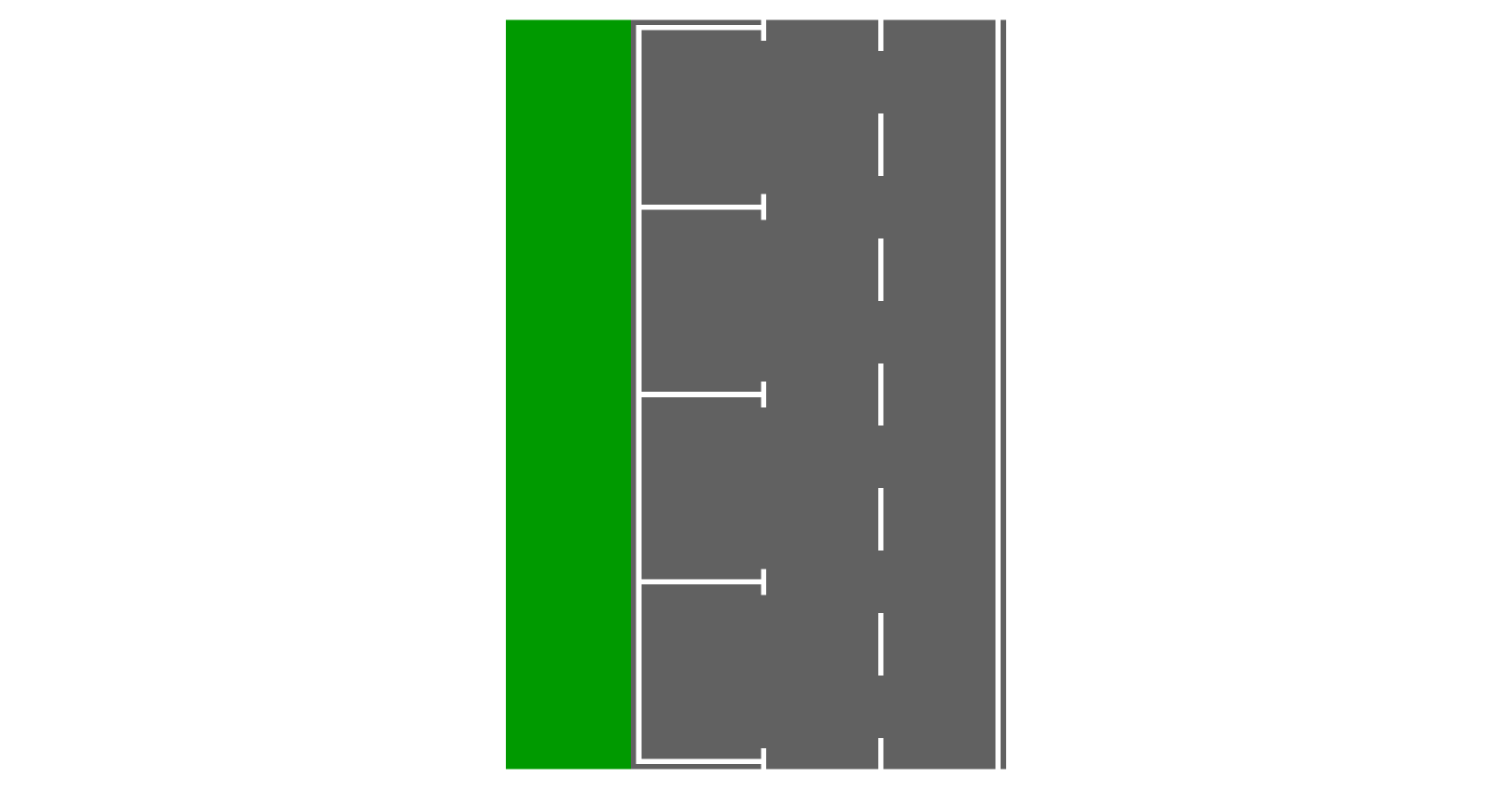}}
	\subfigure[]{\includegraphics[width=0.156\textwidth]{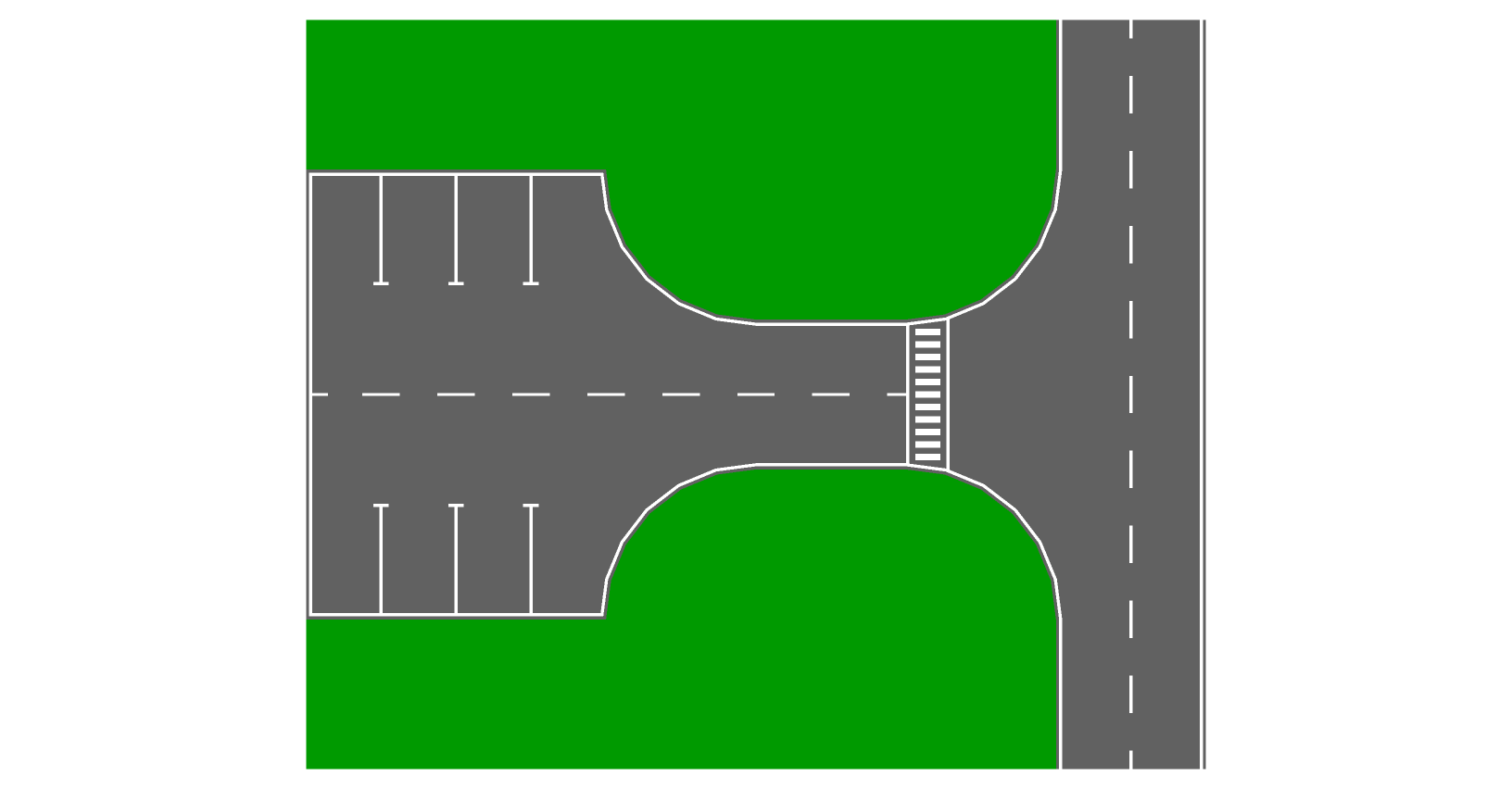}}
	\subfigure[]{\includegraphics[width=0.156\textwidth]{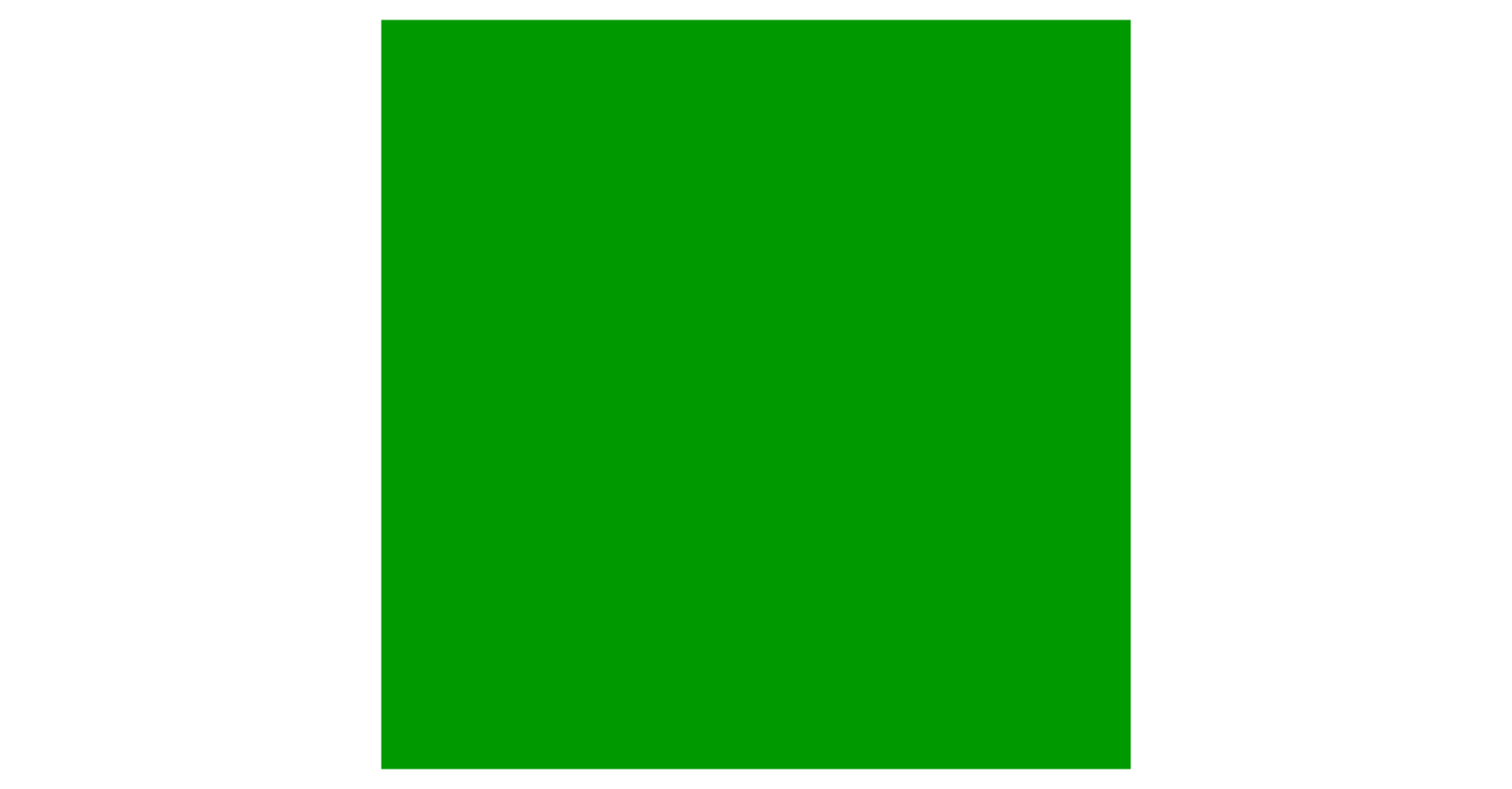}}
	\subfigure[]{\includegraphics[width=0.156\textwidth]{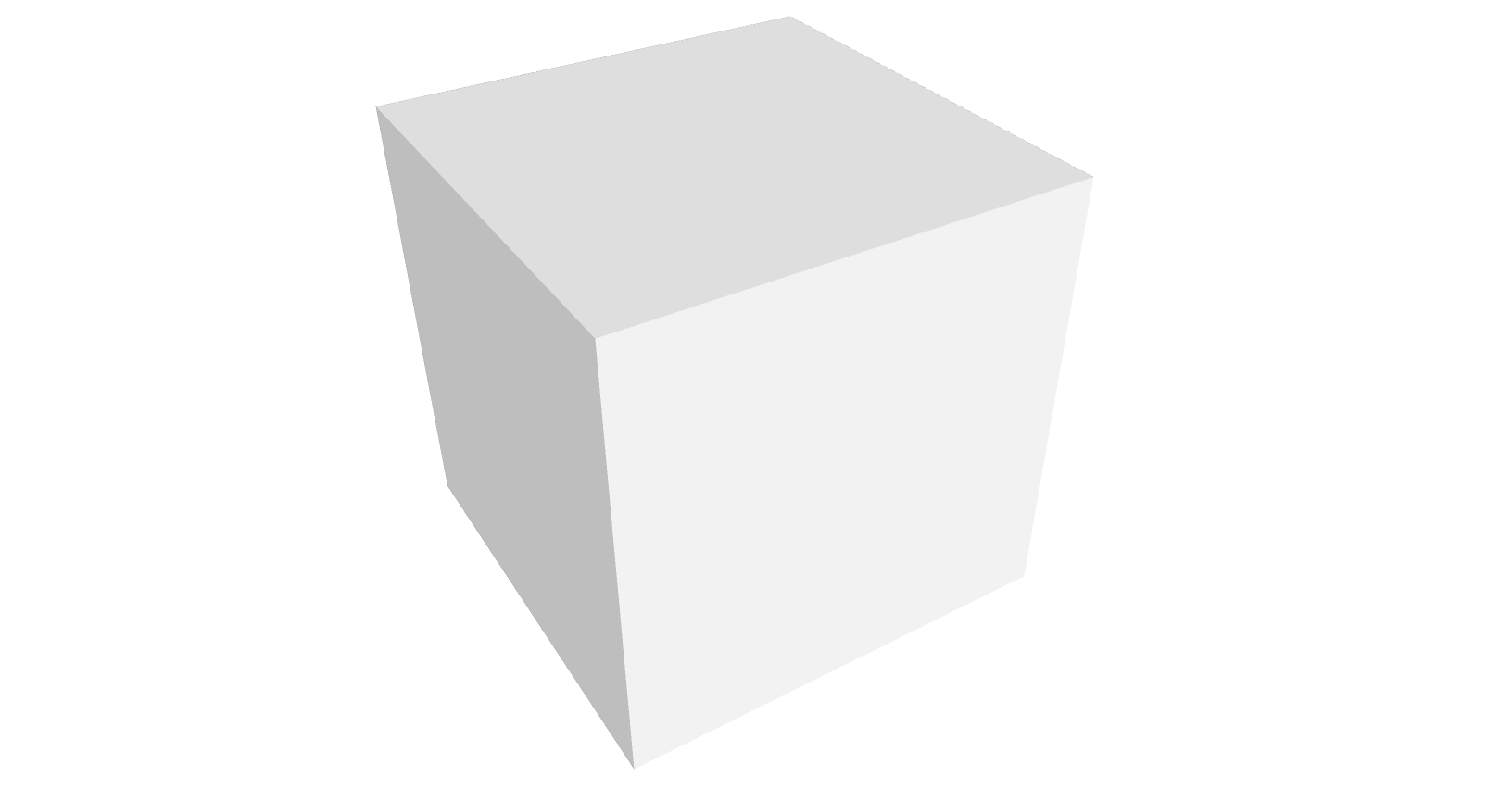}}
	\caption{Currently supported environment modules: (a) dead end, (b) straight road, (c) curved road, (d) 3-way intersection, (e) 4-way intersection, (f) roadside parking, (g) parking lot, (h) lawn, and (i) construction box.}
	\Description{3D mesh models of various environment modules developed using SketchUp.}
	\label{fig:EnvironmentModules}
\end{figure}

\begin{figure}[h]
	\centering
	\includegraphics[width=\linewidth]{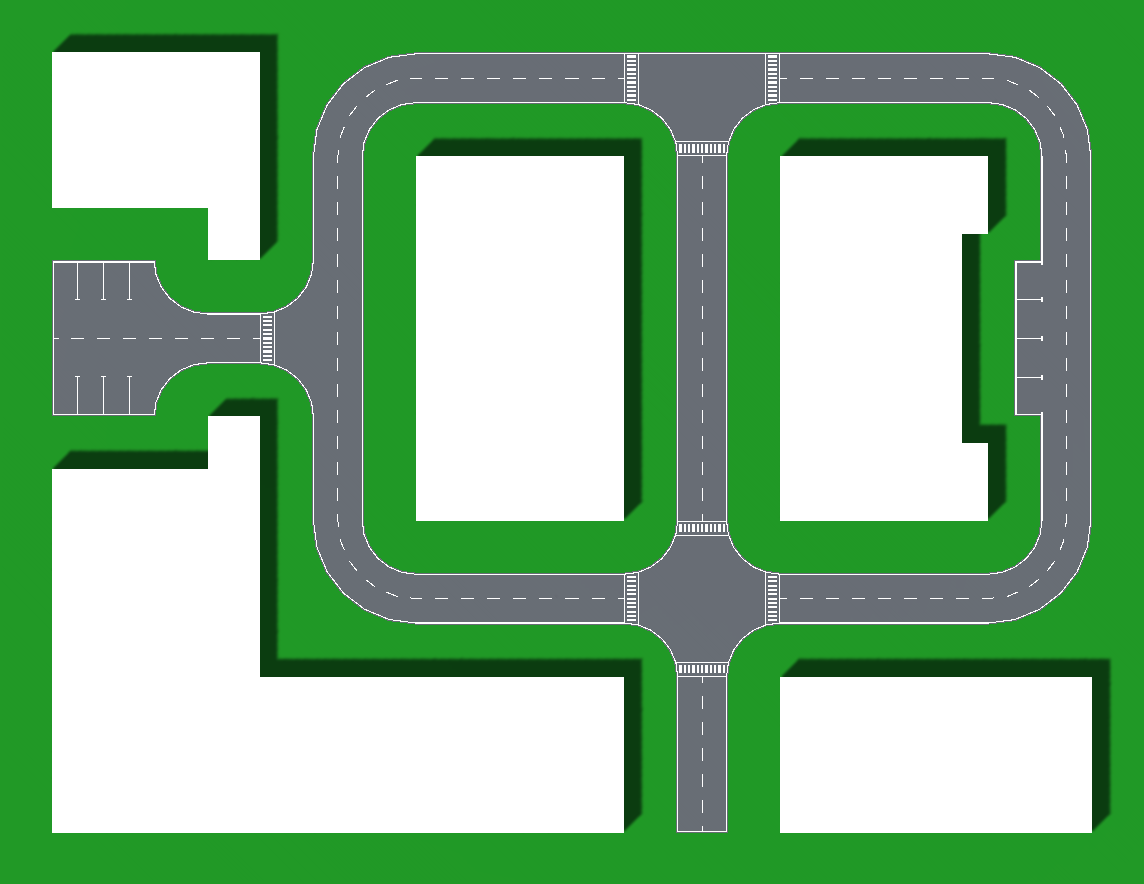}
	\caption{3D architectural map of Tiny Town.}
	\Description{3D architectural map of Tiny Town developed for the AutoDRIVE Simulator, using all the supported environment modules.}
	\label{fig:TinyTown}
\end{figure}

The environment modules, depicted in Figure \ref{fig:EnvironmentModules}, were used to create a small map. The key focus was to encompass all the modules into the map, while also minimizing the space. An additional requirement was to include at least one closed loop path in the map. Figure \ref{fig:TinyTown} depicts \textit{Tiny Town}, a mini-map created by adhering to the aforementioned requirements.

\noindent{\textbf{\\Environment Physics:\\}}

Environment physics simulation includes real-time computation of gravity, friction and other forces, along with collision detection.

\begin{figure}[h]
	\centering
	\includegraphics[width=\linewidth]{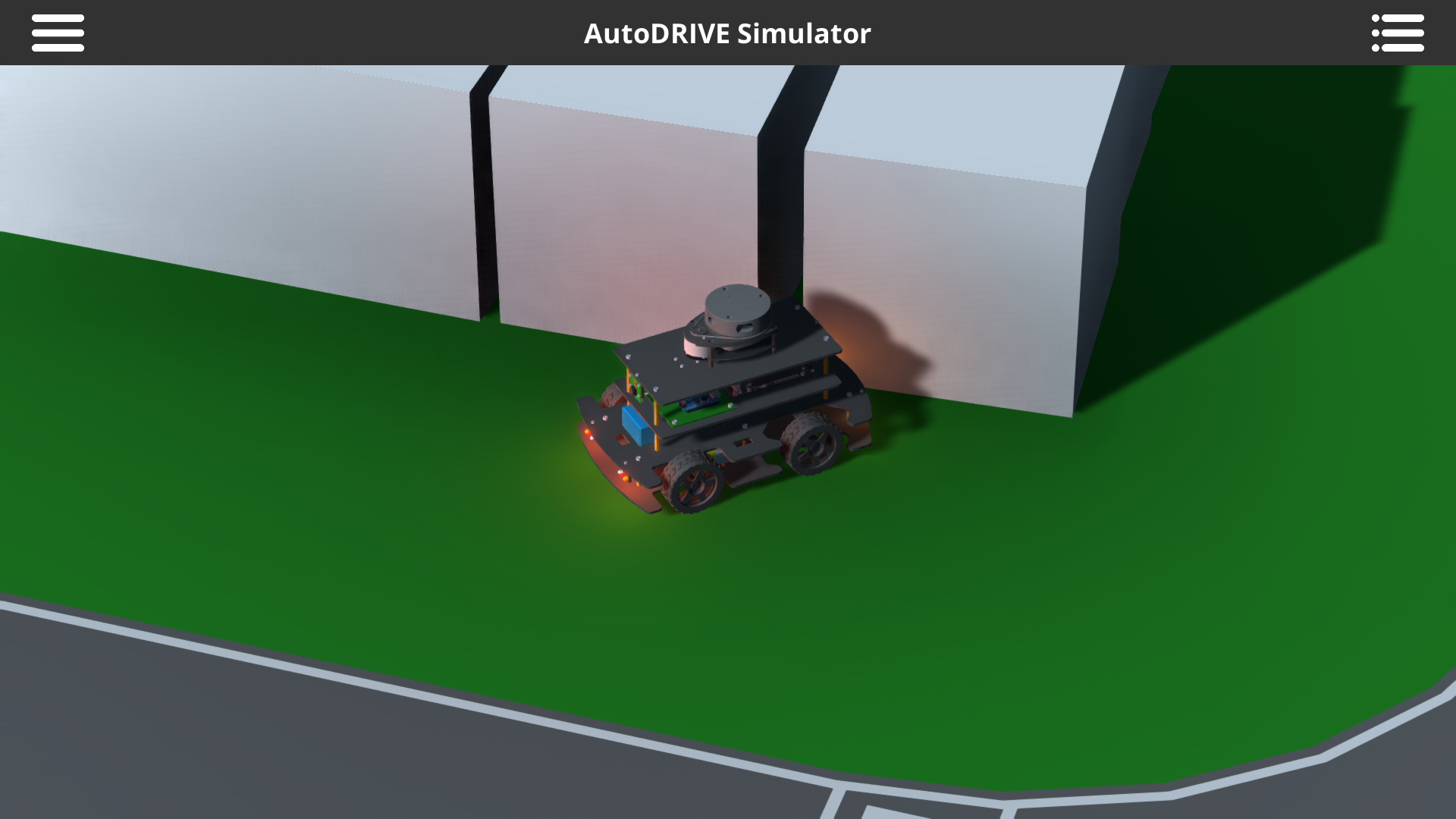}
	\caption{Vehicle collision demonstration.}
	\Description{Vehicle loses control and collides with the construction boxes. The hazard indicators are enabled and vehicle comes to a safe stop. Note the displacement of boxes due to collision.}
	\label{fig:VehicleCollision}
\end{figure}

The 2D environment modules have a box collider attached to them and are defined to be \textit{``static''} in nature. This implies that any interactive forces between the vehicle wheels and these modules shall only affect dynamics of the vehicle. The 3D modules (i.e. construction boxes) also have a box collider attached to them. These modules are defined to be \textit{``dynamic''} objects, which implies that any interaction between the vehicle and these modules shall affect dynamics of both the entities. Furthermore, interaction between these modules is also possible. This is illustrated in Figure \ref{fig:VehicleCollision}.

\noindent{\textbf{Scene Lighting:\\}}

Scene lighting refers to the environmental illumination, which is partially baked (for reducing the computational overhead) and partially computed in real-time (for realistic lighting effects) using ray-casting. This parameter not only affects the on-screen rendering, but also influences visual sensors such as cameras.

\begin{figure}[h]
	\centering
	\subfigure[]{\includegraphics[width=0.23\textwidth]{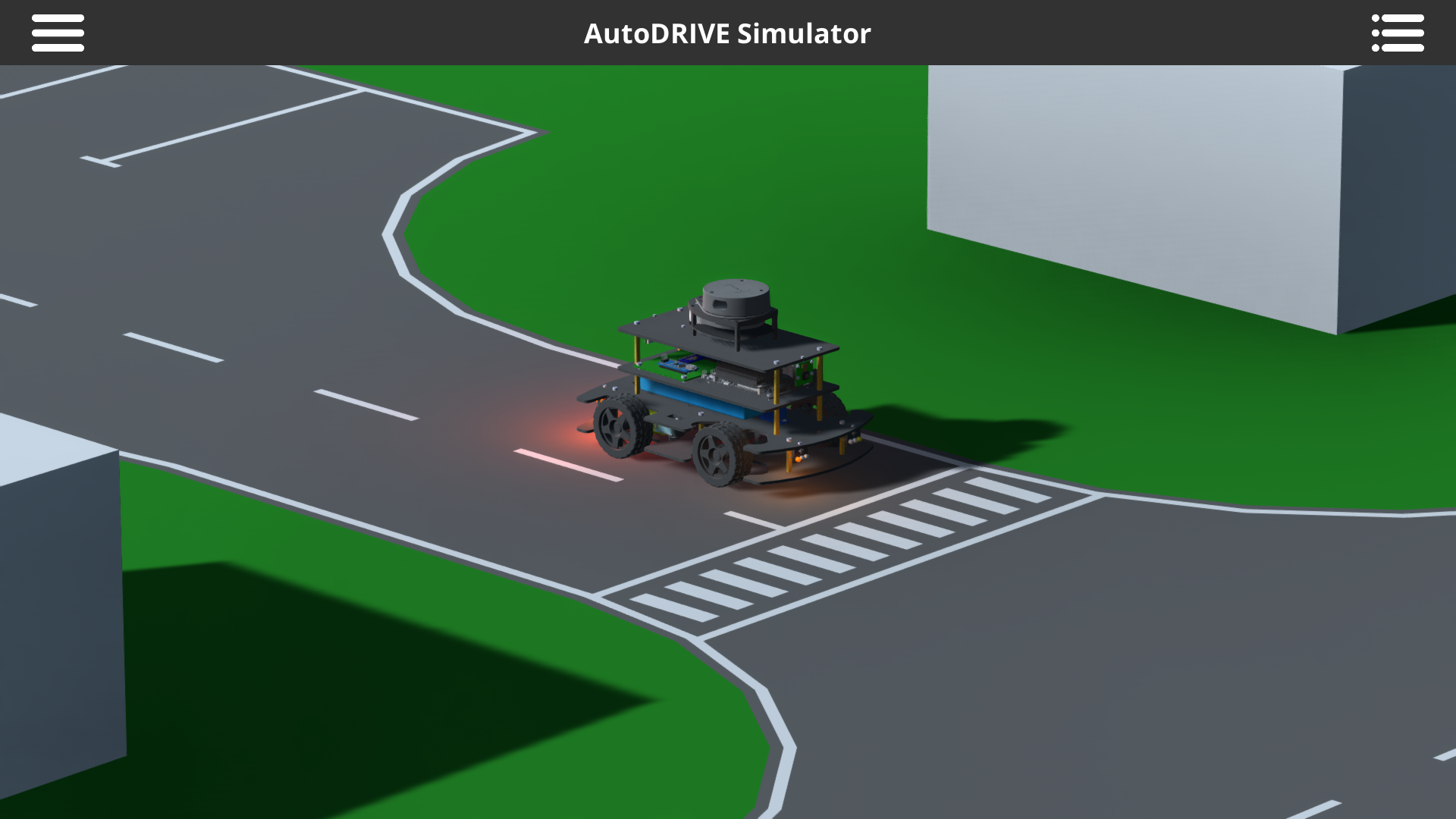}}
	\subfigure[]{\includegraphics[width=0.23\textwidth]{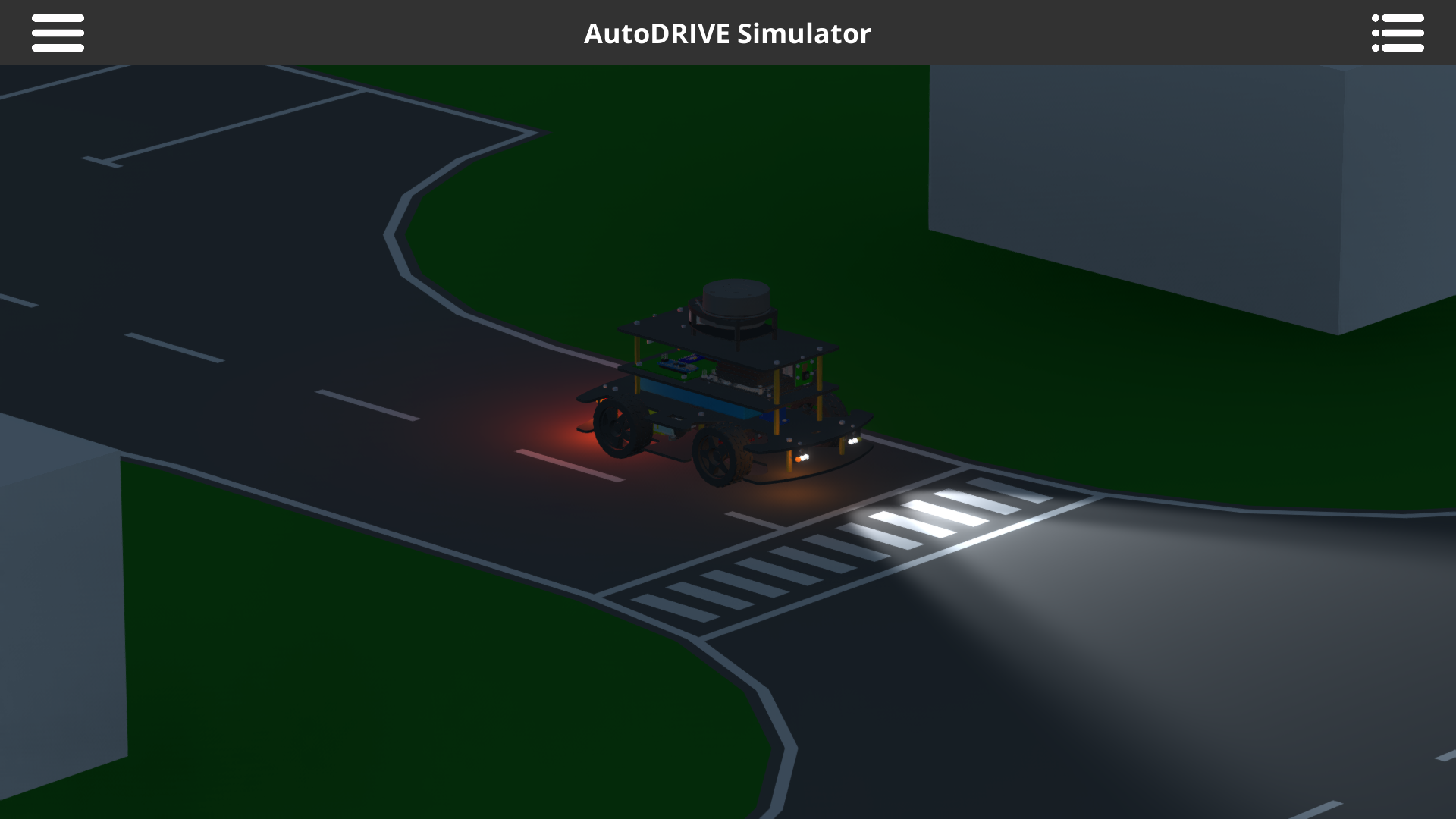}}
	\caption{Scene lighting demonstration: (a) scene light enabled, and (b) scene light disabled.}
	\Description{Vehicle is exiting the parking lot with (a) scene light enabled, and (b) scene light disabled. Note that vehicle headlights are enabled in high-beam mode when the scene light is disabled.}
	\label{fig:SceneLighting}
\end{figure}

The scene light can be enabled (to simulate daylight driving conditions) or disabled (to simulate night driving conditions). This is depicted in Figure \ref{fig:SceneLighting}. While it is not a strict requirement, it is recommended to disable the headlights or use them in low-beam mode while the scene light is enabled; they may be switched to high-beam mode when the scene light is disabled.

\noindent{\textbf{\\Scene Cameras:\\}}

The simulator currently supports three distinct scene cameras.

\begin{figure}[h]
	\centering
	\subfigure[]{\includegraphics[width=0.465\textwidth]{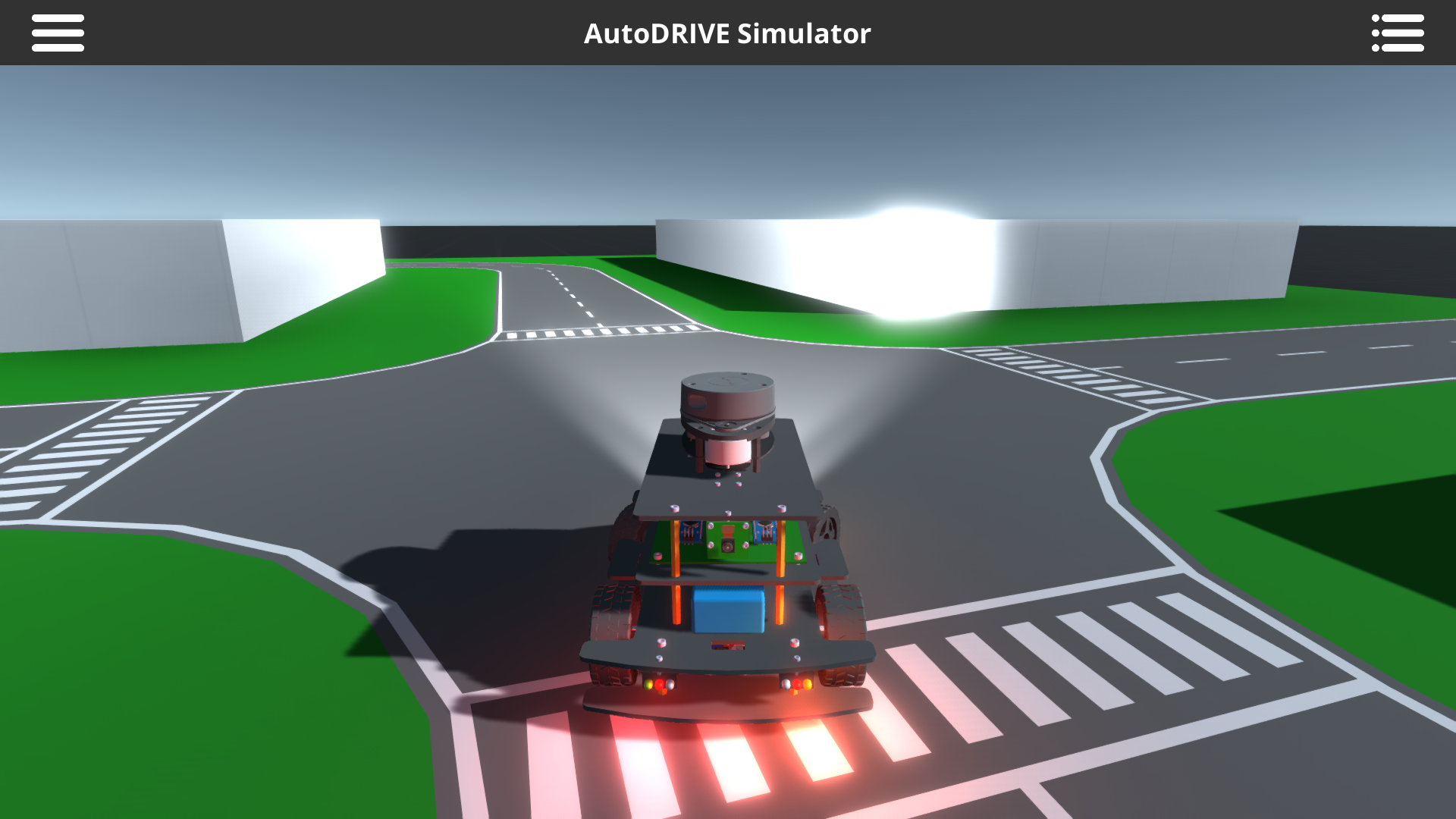}}
	\subfigure[]{\includegraphics[width=0.23\textwidth]{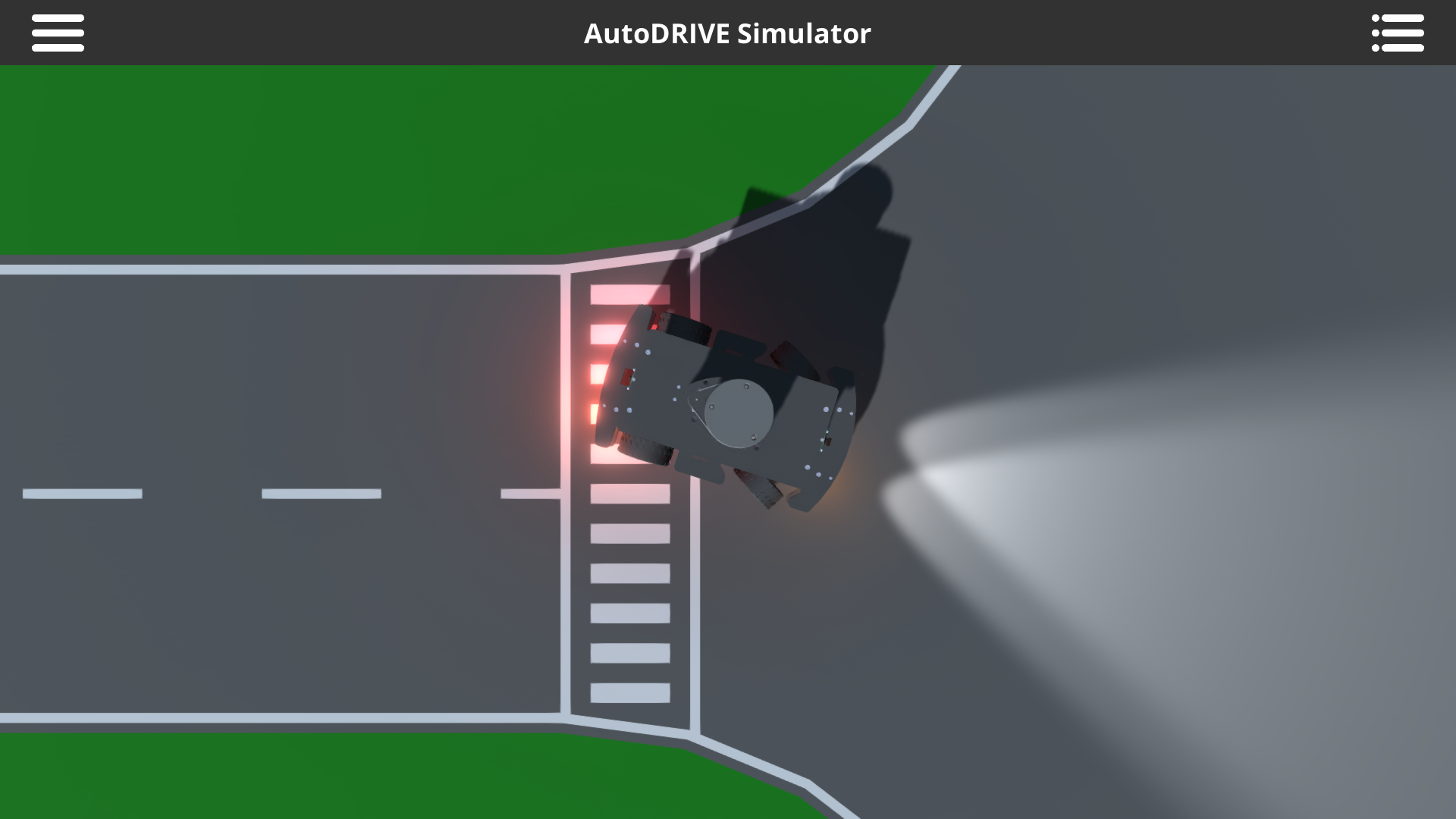}}
	\subfigure[]{\includegraphics[width=0.23\textwidth]{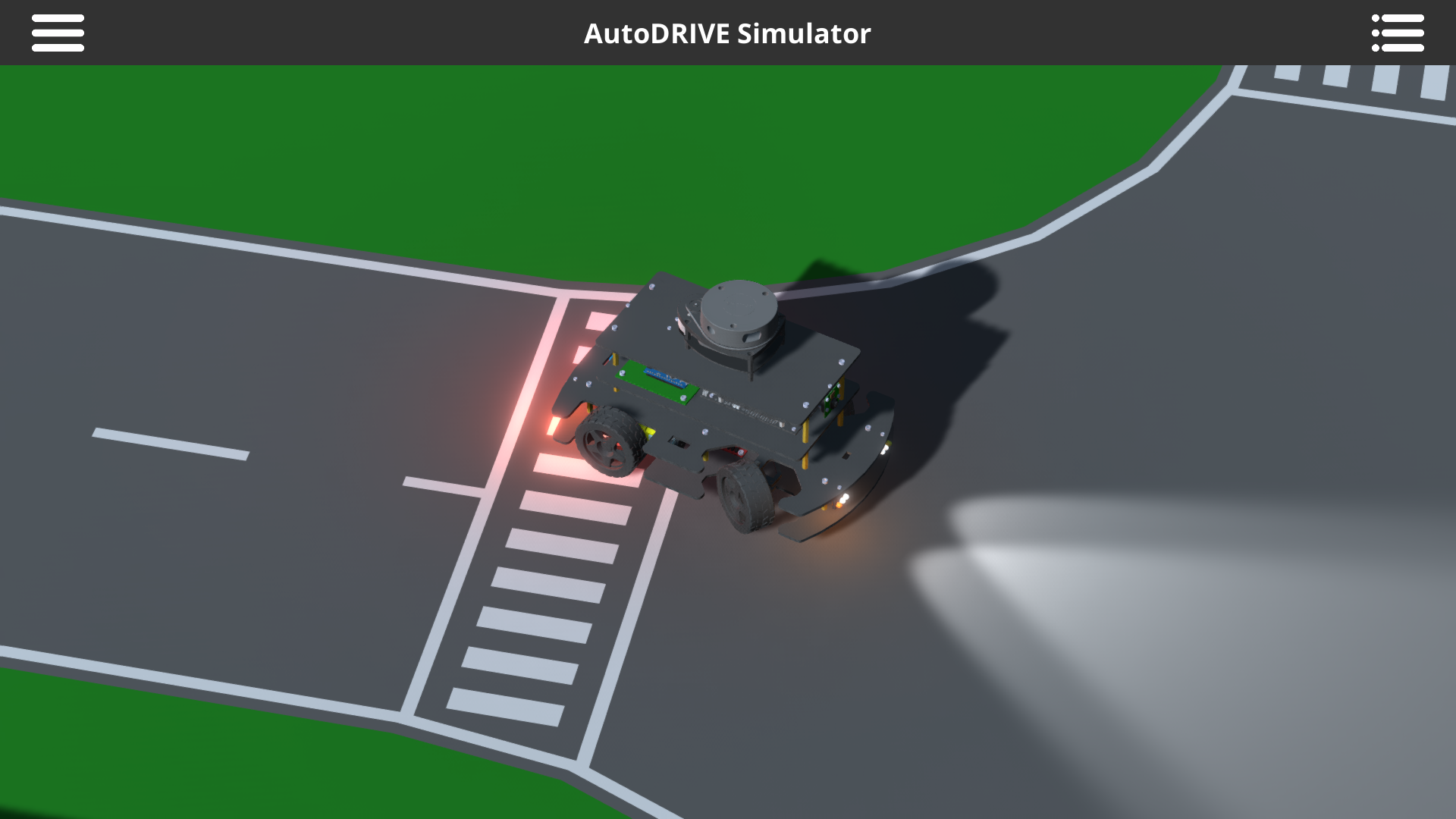}}
	\caption{Scene cameras: (a) Driver's Eye, (b) Bird's Eye, and (c) God's Eye.}
	\Description{Vehicle crossing a 4-way intersection is visualized from (a) Driver's Eye, (b) Bird's Eye, and (c) God's Eye.}
	\label{fig:SceneCameras}
\end{figure}

Each of the three scene cameras (Figure \ref{fig:SceneCameras}) provides a different view:

\begin{itemize}
	
	\item \textbf{Driver's Eye:} This camera continuously follows the vehicle from the rear side and its transform is static w.r.t. the vehicle. This view is convenient for teleoperating the vehicle manually or to follow it, in general.
	
	\item \textbf{Bird's Eye:} This camera continuously tracks the vehicle from a top view (without rotating) and has a provision to zoom in or out as required in order to give the user a better visualization of the vehicle within the environment. This view is convenient for driving the vehicle manually (especially in reverse gear) or for scenic visualization/recording in autonomous driving mode.
	
	\item \textbf{God's Eye:} This camera continuously tracks the vehicle while automatically adjusting its focus and field of view for efficient visualization. This view is especially useful for scenic visualization/recording in autonomous driving mode.
	
\end{itemize}

\subsection{Communication Bridge}

The communication bridge is implemented using WebSocket, which provides full-duplex communication channels over a single TCP connection. Another peculiar advantage of WebSocket is that it allows event-driven responses, which significantly enhances the bridge functionality. Finally, WebSocket also minimizes overhead per message, thereby making it an efficient protocol for our application.

The bridge parameters are reconfigurable, thereby allowing users to enter specific IP address and port number in order to establish a server-client communication between their scripts (server) and the simulator (client). This allows local as well as distributed computing, meaning users can run the simulator and scripts on the same machine (configured as default bridge parameters \texttt{127.0.0.1:4567}) or on different machines connected through a common network (LAN/WLAN) by specifying custom bridge parameters. This expands the computation limits of the simulation system.

\subsection{User Interface}

AutoDRIVE Simulator aims at providing supreme user experience. There are two fundamental components of the user interface viz. graphical user interface (GUI) and hardware input methods.

\noindent{\textbf{\\Graphical User Interface:\\}}

\begin{figure}[h]
	\centering
	\includegraphics[width=\linewidth]{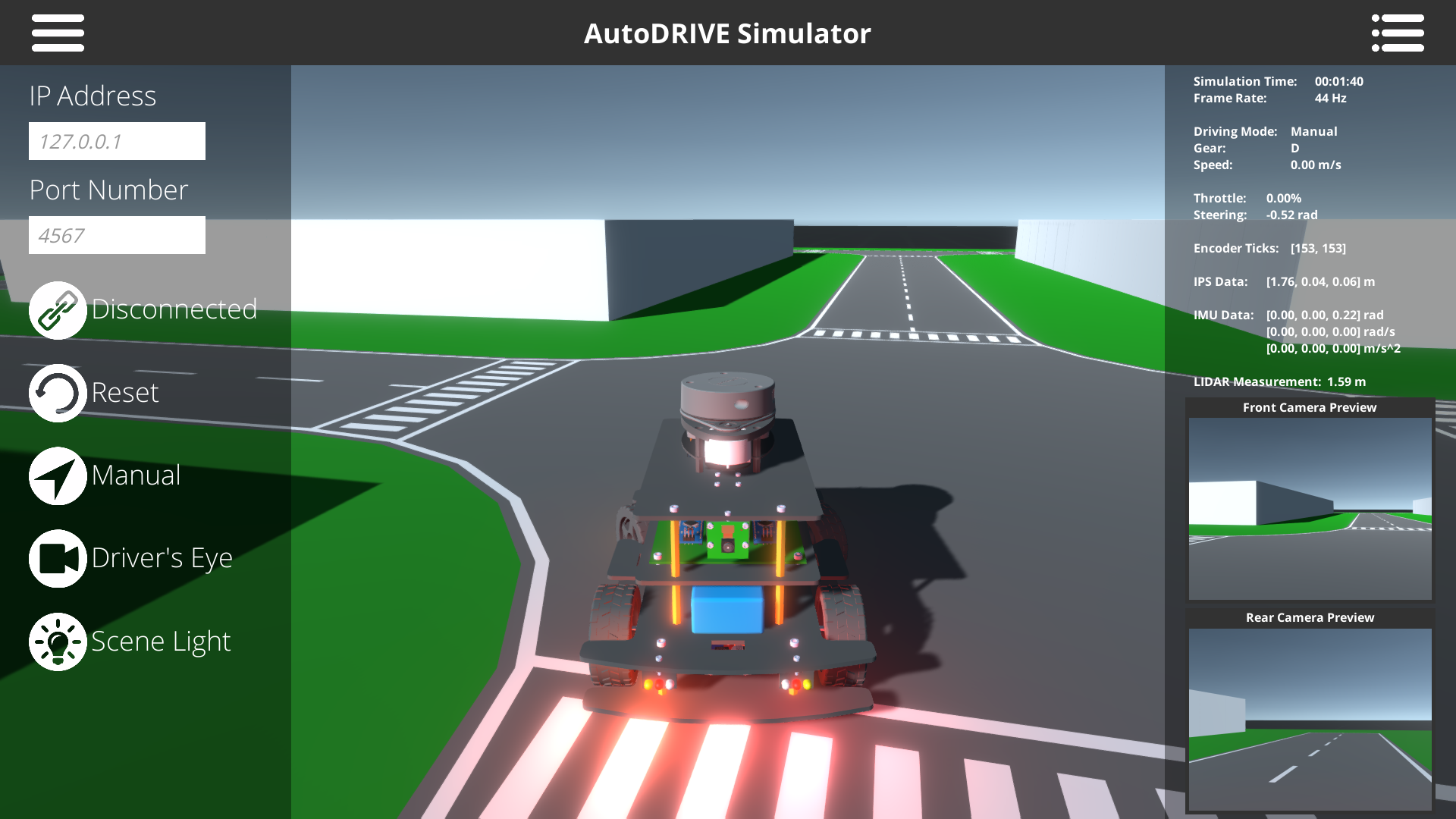}
	\caption{Graphical User Interface of AutoDRIVE Simulator.}
	\Description{Screenshot of AutoDRIVE Simulator with both GUI panels enabled. The communication bridge is disconnected, vehicle is in manual driving mode, scene camera is switched to Driver's Eye, and scene light is enabled. 1 minute and 40 seconds have elapsed since start of simulation and the simulator is rendering at 44 Hz.}
	\label{fig:GUI}
\end{figure}

The GUI consists of a toolbar encompassing two panels, namely Menu and Heads-Up Display (HUD). Both panels can be enabled or disabled using buttons provided on the toolbar. Figure \ref{fig:GUI} illustrates both GUI panels being enabled -- the left one is Menu panel, while the right one is HUD panel.

\noindent{\textit{\textbf{\\Menu Panel:}}}

\begin{itemize}
	
	\item \textbf{IP Address Field:} Input field to specify IP address for the communication bridge (default is \texttt{127.0.0.1}, i.e. standard address for IPv4 loopback traffic).
	
	\item \textbf{Port Number Field:} Input field to specify port number for the communication bridge (default is \texttt{4567}).
	
	\item \textbf{Connection Button:} Button to establish connection with the server (the button is disabled once the connection is established). The status of bridge connection (i.e. Connected or Disconnected) is displayed besides this button.
	
	\item \textbf{Reset Button:} Button to reset the simulator to initial conditions.
	
	\item \textbf{Driving Mode Button:} Button to toggle the driving mode of the vehicle between Manual and Autonomous (default is Manual). The selected driving mode is displayed besides this button.
	
	\item \textbf{Camera Button:} Button to toggle the scene camera view between Driver’s Eye, Bird’s Eye and God’s Eye (default is Driver’s Eye). The selected view is displayed besides this button.
	
	\item \textbf{Scene Light Button:} Button to enable/disable the scene light (default is enabled).
	
\end{itemize}

\noindent{\textit{\textbf{HUD Panel:}}}

\begin{itemize}
	
	\item \textbf{Simulation Time:} The time (HH:MM:SS) since start of the simulation. Reset button resets the simulation time.
	
	\item \textbf{Frame Rate:} Running average of the FPS value (Hz).
	
	\item \textbf{Driving Mode:} Driving mode of the ego-vehicle (Manual or Autonomous).
	
	\item \textbf{Gear:} Drive direction of the vehicle; either Drive (D) or Reverse (R).
	
	\item \textbf{Speed:} Magnitude of forward velocity of the vehicle (m/s).
	
	\item \textbf{Throttle:} Instantaneous throttle input of the vehicle (\%).
	
	\item \textbf{Steering:} Instantaneous steering angle of the vehicle (rad).
	
	\item \textbf{Encoder Ticks:} Ticks (counts) of the left and right incremental encoders of the vehicle represented using a 1D array of 2 elements \texttt{[left\_ticks, right\_ticks]}.
	
	\item \textbf{IPS Data:} Position (m) of vehicle within the map represented using a vector \texttt{[x, y, z]}.
	
	\item \textbf{IMU Data:} Orientation \texttt{[x, y, z] rad}, angular velocity \texttt{[x, y, z] rad/s}, and linear acceleration \texttt{[x, y, z] m/s\textsuperscript{2}} of the ego-vehicle about its local axes.
	
	\item \textbf{LIDAR Measurement:} Instantaneous ranging measurement (m) of the LIDAR.
	
	\item \textbf{Camera Previews:} Instantaneous raw images from the front and rear-view cameras of the vehicle.
	
\end{itemize}

\noindent{\textbf{Hardware User Interface:\\}}

The simulator currently accepts input from hardware interfaces including keyboard and mouse. In addition to interacting with the GUI, the hardware inputs also serve as key control elements for the simulated vehicle. Table \ref{tab:HardwareUserInterface} summarizes the hardware inputs and their respective functionalities.

\begin{table}[h]
	\caption{Hardware User Interface of AutoDRIVE Simulator}
	\label{tab:HardwareUserInterface}
	\begin{tabular}{lll}
		\toprule
		\textbf{Hardware} & \textbf{User Input}             & \textbf{Function}			\\
		\midrule
		Keyboard          & W/Up arrow                      & Drive vehicle forward		\\
		Keyboard          & S/Down arrow                    & Drive vehicle reverse		\\
		Keyboard          & A/Left arrow                    & Steer vehicle left		\\
		Keyboard          & D/Right arrow                   & Steer vehicle right		\\
		Keyboard          & G                               & Headlights (low-beam)		\\
		Keyboard          & H                               & Headlights (high-beam)	\\
		Keyboard          & L                               & Left indicators			\\
		Keyboard          & R                               & Right indicators			\\
		Keyboard          & E                               & Hazard indicators			\\
		Keyboard          & A-Z, 0-9, spl. char.			& Input Data through GUI	\\
		Mouse             & Click                           & Interact with GUI			\\
		Mouse             & Scroll                          & Zoom (Bird’s Eye view)	\\
		\bottomrule
	\end{tabular}
\end{table}

\section{Development Framework}

The development framework is a collection of tools that enable users to exploit the AutoDRIVE Simulator for rapid and flexible development of autonomy algorithms. It supports local as well as distributed computing and is compatible with ROS, while also offering a direct scripting support for Python and C++.

\subsection{ROS Package}

\begin{figure}[h]
	\centering
	\includegraphics[width=\linewidth]{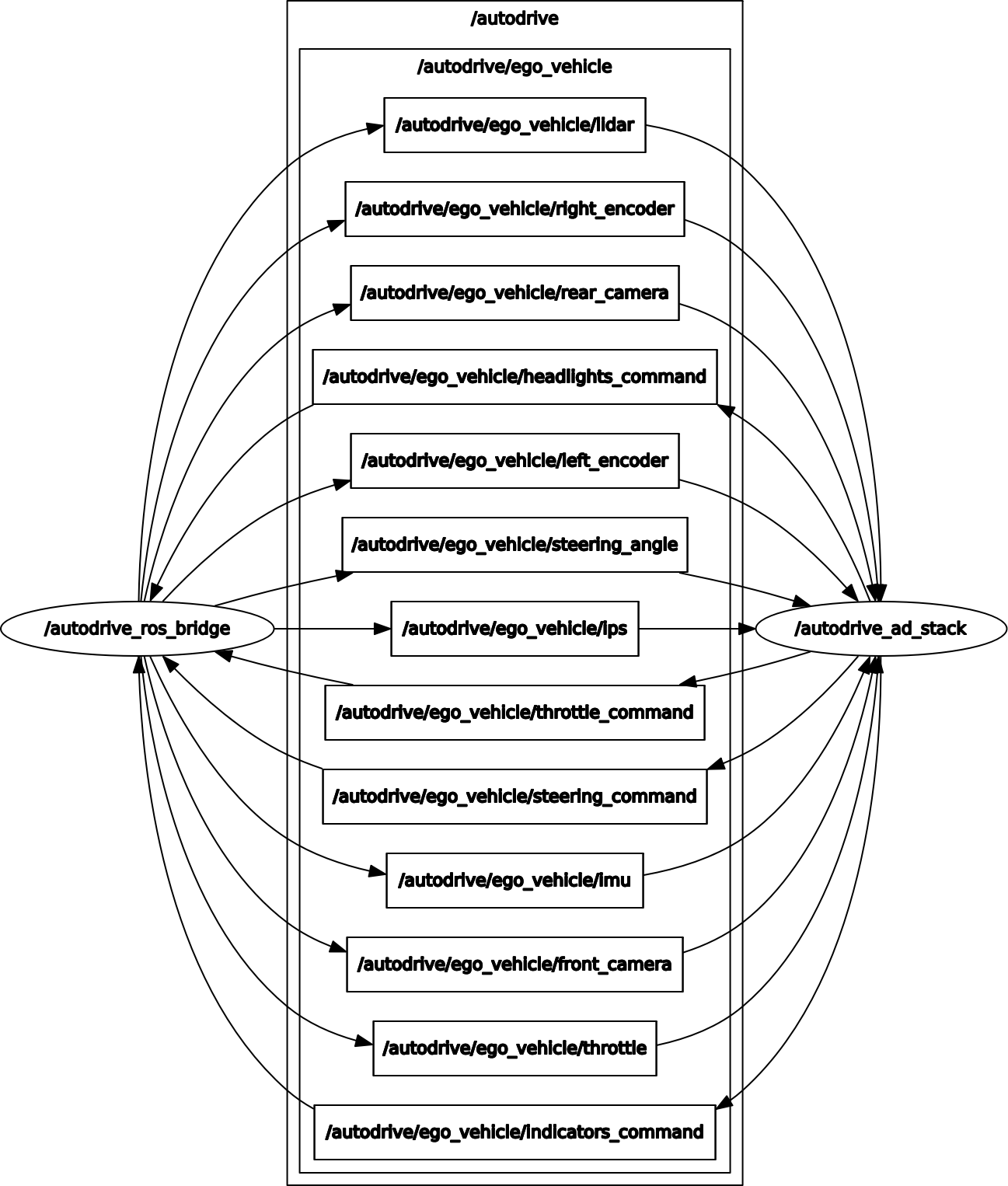}
	\caption{ROS computation graph captured after interfacing AutoDRIVE Simulator with ROS\protect\footnotemark.}
	\Description{Screenshot of ROS computation graph captured after interfacing the AutoDRIVE Simulator (running on a Windows machine) with ROS (running on a separate Ubuntu machine) by executing autodrive.launch file from the autodrive ROS package.}
	\label{fig:RQTGraph}
\end{figure}
\footnotetext{Please note that \texttt{/autodrive\_ad\_stack} node is developed for representational purpose and that users can implement autonomous driving software stack as they please.}

The ROS package developed as a part of this research project aims at modular development of autonomous driving software stack. The package comprises of \texttt{/autodrive\_ros\_bridge} node, which handles bilateral communication with the simulator. The node receives sensor data from the simulator and publishes them as standard sensor messages on appropriate topics. The node also subscribes to the final control commands generated by the autonomous driving software stack for driving the vehicle (in autonomous driving mode) and transmits those control commands to the simulator. Figure \ref{fig:RQTGraph} depicts a sample ROS computation graph. As an icing on the cake, the ROS package also includes a pre-configured ROS Visualization (RViz) file in order to visualize the vehicle state and supported sensor readings on the server-side (Figure \ref{fig:Teaser}).

\subsection{Scripting APIs}

AutoDRIVE Simulator also supports general purpose scripting APIs for Python and C++ that were developed as a part of this research project. These interfaces can be used to directly communicate with the simulator without ROS (or any other middleware) as an intermediary. The scripting interfaces currently support receiving sensor data from the simulator and transmitting control commands back to it. These interfaces shall prove to be extremely efficient during the prototyping phase of autonomy algorithms as they allow direct deployment.

\section{Conclusion}

In this work, we introduced AutoDRIVE Simulator, a simulation ecosystem for scaled autonomous vehicles and related applications. We also disclosed some of the prominent components of this simulation system as well as some key features that it has to offer in order to accelerate education and research in the field of autonomous systems, autonomous driving in particular.

This simulator already has a lot to offer so as to boost the iterative development-deployment cycle for students and researchers. However, further enhancement shall only make it more useful for others in the field. This simulator was developed as a research project as a part of India Connect@NTU Research Internship Programme, 2020 within the stipulated duration of two months and there lies a great scope for further development of this work. We shall therefore actively continue contributing to this research project with a primary aim of addressing the needs of students, educators and researchers.

\section*{Supplemental Material}

\noindent{\textbf{Source Code:}} \url{https://github.com/Tinker-Twins/AutoDRIVE}

\noindent{\textbf{Video:}} \url{https://youtu.be/i7R79jwnqlg}

\bibliographystyle{ACM-Reference-Format}
\bibliography{References}

\end{document}